\documentclass[10pt,twocolumn,letterpaper]{article}

\usepackage[pagenumbers]{iccv} 

\usepackage[accsupp]{axessibility}  

\definecolor{iccvblue}{rgb}{0.21,0.49,0.74}
\usepackage[pagebackref,breaklinks,colorlinks,allcolors=iccvblue]{hyperref}

\usepackage{amsmath,amsfonts,bm}

\def\eqref#1{equation~\ref{#1}}

\def\1{\bm{1}}

\DeclareMathAlphabet{\mathsfit}{\encodingdefault}{\sfdefault}{m}{sl}
\SetMathAlphabet{\mathsfit}{bold}{\encodingdefault}{\sfdefault}{bx}{n}

\usepackage{url}
\usepackage[capitalize]{cleveref}
\usepackage{graphicx}
\usepackage{booktabs}
\usepackage{multirow}
\usepackage[accsupp]{axessibility}
\def\paperID{9081} 
\def\confName{ICCV}
\def\confYear{2025}

\newcommand{\method}{\textsc{CAPTURe}}
\newcommand{\baseline}{\textsc{CountGD}}
\newcommand{\inpaint}{Inpainting Pipeline}

\newcommand{\synthetic}{\method{}$^\text{synthetic}$}
\newcommand{\real}{\method{}$^\text{real}$}

\usepackage[framemethod=TikZ]{mdframed}
\definecolor{OurColor}{HTML}{36aa70}
\definecolor{UserExampleBg}{HTML}{ffffff}
\definecolor{UserExampleTitle}{HTML}{545f7f}
\newmdenv[
    roundcorner=5pt,
    backgroundcolor=UserExampleBg,
    linecolor=UserExampleTitle,
    outerlinewidth=0.5pt,
    frametitlebackgroundcolor=UserExampleTitle,
    frametitlefont={\bfseries\color{white}},
]{user_example}

\AtBeginEnvironment{user_example}

\newcommand{\myparagraph}[1]{\vspace{0.75em}\noindent\textbf{#1}\hspace{0.5em}}

\title{\method{}: Evaluating Spatial Reasoning in Vision Language Models\\ via Occluded Object Counting}

\author{
Atin Pothiraj
\qquad
Elias Stengel-Eskin
\qquad
Jaemin Cho
\qquad
Mohit Bansal\\
UNC Chapel Hill\\
{\tt\small \qquad \{atin, esteng, jmincho, mbansal\}@cs.unc.edu}\\
}

\begin{document}
\maketitle
\begin{abstract}
Recognizing and reasoning about occluded (partially or fully hidden) 
objects is vital to understanding visual scenes, as occlusions frequently occur in real-world environments and act as obstacles for spatial comprehension. 
To test models' ability to reason about multiple occluded objects, we introduce a novel task, \textbf{C}ounting \textbf{A}modally for \textbf{P}atterns \textbf{T}hrough \textbf{U}nseen \textbf{RE}gions (\method{}), which requires a model to count objects arranged in a pattern by inferring how the pattern continues behind an occluder (an object which blocks parts of the scene).
\method{} requires both recognizing visual patterns and reasoning, making it a useful testbed for evaluating vision-language models (VLMs) on whether they understand occluded patterns and possess spatial understanding skills. 
By requiring models to reason about occluded objects, \method{} also tests VLMs' ability to form world models that would allow them to fill in missing information. 
\method{} consists of two parts:
(1) \real{}, with manually filtered images of real objects in patterns 
and (2) \synthetic{}, a controlled diagnostic with generated patterned images. 
We evaluate four strong VLMs (GPT-4o, Intern-VL2, Molmo, and Qwen2-VL) on \method{}, finding that models struggle to count on both occluded and unoccluded patterns. 
Crucially, we find that models perform worse with occlusion, suggesting that VLMs are also deficient in inferring unseen spatial relationships: even the strongest VLMs like GPT-4o fail to count with occlusion. 
In contrast, we find that humans achieve very little error on \method{}. 
We also find that providing auxiliary information of occluded object locations increases performance, underscoring that the model error comes both from an inability to handle occlusion as well as difficulty in counting in images.\footnote{Code and data: \href{https://github.com/atinpothiraj/CAPTURe}{https://github.com/atinpothiraj/CAPTURe}}
\end{abstract}

\vspace{-0.75em}
\section{Introduction}
\label{sec:intro}
\vspace{-0.25em}

\begin{figure}[t]
    \centering
    \includegraphics[width=\linewidth]{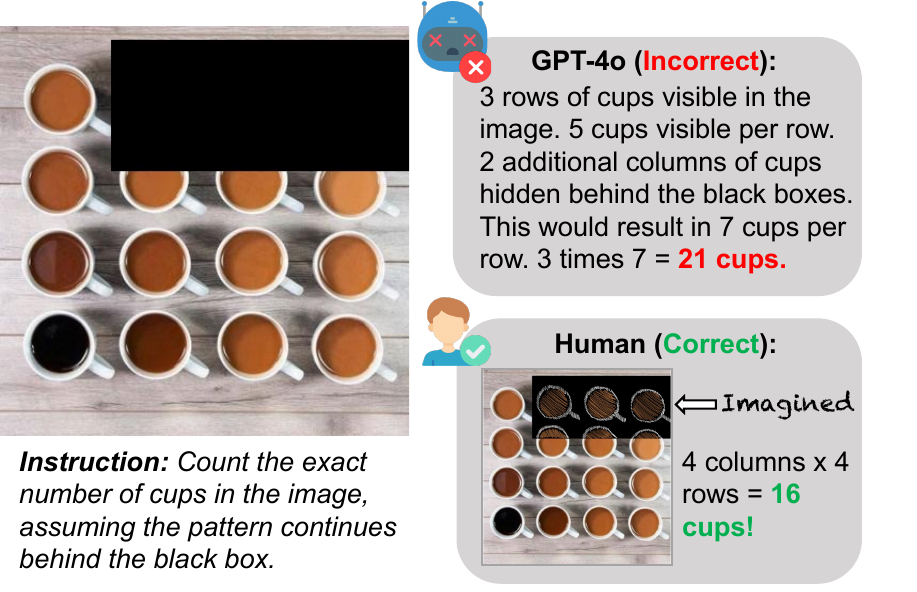}
    \caption{\method{} example with an output from GPT-4o. 
    While people can easily infer the missing number of cups and correctly reason over occluded patterns, models generally struggle to reason over these occluded scenes.}
    \vspace{-1em}
    \label{fig:fig1}
\end{figure}

Inferring what lies behind different objects in occluded scenes is crucial for human perception, as it allows us to maintain a coherent understanding of our environment even when parts are hidden.
The human visual system accomplishes this by integrating past experiences, context, and sensory inputs to reconstruct incomplete scenes~\cite{otsuka2006development, olson2004neuronal, wynn1990children,gestalt_perception}.
Meanwhile, recent advancements in vision-language models (VLMs) -- especially in terms of visual and spatial reasoning -- raise the question of whether these systems can perform similar inferential tasks.
One way of measuring such capabilities is through amodal completion -- the task of inferring the invisible parts of partially occluded objects; here, vision-only models are typically evaluated via dense prediction tasks like object segmentation and image inpainting \cite{ao2023image}.
However, this format is not well-suited for assessing VLMs, whose outputs consist of text tokens rather than pixel-level predictions. 
This raises a critical question: How can we quantify the ability of VLMs to form spatial world modeling~\cite{ha2018recurrent} in the presence of occlusion?

To address this, we introduce \textbf{\method{}}, \textbf{C}ounting \textbf{A}modally for \textbf{P}atterns \textbf{T}hrough \textbf{U}nseen \textbf{RE}gions,
a novel benchmark that tests a VLM's world modeling and spatial reasoning abilities through the task of \textit{amodal counting}, where models are prompted to count occluded objects by amodally completing a pattern.
\method{} focuses on counting as it provides an objective and easy-to-verify output by comparing predicted counts with ground truth values.
Moreover, patterned objects appear in various real-world domains, especially in man-made environments like parking lots, cities, and warehouses, where counting objects is often required. 
\cref{fig:fig1} illustrates the \method{} task. 
We show a VLM an image where objects are placed in a regular pattern (e.g., a 4x4 grid) with some objects occluded, and ask the model to count the total number of objects in the image \textit{assuming that the pattern continues behind the occlusion}.
The task requires \emph{handling occlusion}, \emph{pattern recognition}, and \emph{counting} skills that exist in humans from a fairly young age~\citep{otsuka2006development, olson2004neuronal, wynn1990children}, thus humans can easily answer such questions -- indeed, we find that people can complete \method{} tasks with almost no error. 

\begin{figure*}[h]
    \centering
    \includegraphics[width=\linewidth]{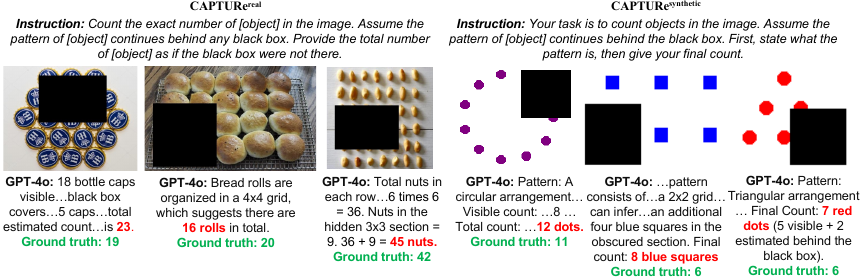}
    \caption{Example images with GPT-4o responses to \real{} and \synthetic{} occluded splits. 
    }
    \label{fig:examples}
\end{figure*}

\method{} consists of two subsets: \real{} and \synthetic{}.
As shown in~\cref{fig:examples},
\real{} contains real-world images and tests the ability of models to perform amodal counting in naturalistic contexts, while \synthetic{} allows us to analyze specific factors by controlling different variables like color, shape, and number of objects.
All images in \method{} contain a pattern of objects and a manually annotated occluding black box covering some objects. 
\real{} contains 924 images with a diverse range of settings and objects, covering 92 different object types, while \synthetic{} contains 1250 images across multiple attribute classes.

By combining vision encoders with large language models (LLMs), VLMs have the potential to reason in a zero-shot way about visual inputs.
To put this ability to the test and measure VLMs' ability to reason about missing visual information, we evaluate four strong recent VLMs (GPT-4o, InternVL2, Molmo, and Qwen2VL) on \method{}. 
Our experiment results (\cref{sec:result}) show that models generally struggle with the multiple aspects of the task, with high error rates on both \real{} and \synthetic{} for occluded and unoccluded images. 
In contrast, we find that humans can perform the task easily: whereas model performance deteriorates as more objects in images are occluded, humans complete the task almost perfectly.
We also compare VLMs to a vision-only model trained to count visible objects; while this model generally outperforms VLMs, its error is directly tied to the number of occluded objects -- the more objects are occluded, the higher its error will be. 

By objectively measuring VLMs' spatial reasoning capabilities under occlusion, \method{} highlights an unexpected weakness in VLMs. 
We analyze this weakness by providing the model with additional clues and information. 
Specifically, we test to what degree the VLMs' failure stems from an inability to integrate visual information by providing it with a text-based representation of the visible objects in the image in the form of object coordinates; here, VLMs perform substantially better, indicating that their poor performance on \method{} stems partly from an inability to count objects in images, rather than an inability to count more generally.
Our findings align with previous work, which similarly finds that VLMs struggle to count in images \cite{qharabagh2024lvlm,li2024naturalbench,wang2025seeing}.
We also test the degree to which VLM errors stem from an inability to form a world model by providing it with auxiliary information
(the coordinates of the occluded objects in text,
or inpainting the occluded regions). 
We find that VLMs perform substantially better with this auxiliary information, suggesting that VLMs are partly limited by their inability to imagine the missing visual information. 
Addressing these gaps is critical for VLMs to function effectively in real-world scenarios, where visual reasoning often involves occlusions -- whether counting stadium seats, components on production lines, or buildings in neighborhoods.
We hope that our work will foster future research on improving the world modeling capabilities of VLMs.

\section{\method{}}
\label{sec:method}

\subsection{Task Overview}
\paragraph{Input/output formulation.}
\method{} tests VLMs on occlusion reasoning, pattern recognition, and counting of both visible and occluded objects.
VLMs already achieve high accuracy in classifying single, occluded objects \cite{kassaw2024deep}. 
Thus, we also argue that VLMs have the potential to perform well on \method{}'s challenging task because their proficiency in handling occlusion ought to enable them to recognize occluded objects and reason accordingly. 
All images in \method{} contain a pattern. This makes the task solvable for models and people -- if the objects were not placed in a pattern, it would be unreasonable to expect models to infer the position of the occluded objects. 
For example, given an image of a random pile of coins with a region occluded, it is not easy to infer whether the occluded region contains no coins or contains roughly the same amount as the rest of the pile. 
For this task, the patterns considered are all regular and fairly small, e.g. grids, circles, triangles, and other regular shapes
-- see \cref{fig:examples} for further examples.
The last step of \method{} is counting, asking the model to provide an objectively measurable output. 
In addition to VLMs, we also test \baseline{} \citep{amini2024countgd}, a state-of-the-art object detection-based counting method, finding that it fails to account for the occluded scenario, as its training entails solely predicting the visible, unoccluded objects in the image. 

\myparagraph{Metric.}
We use symmetric mean percent error (sMAPE) as the primary metric. sMAPE is given by:

\begin{equation}
\text{sMAPE} = 100 \cdot \frac{1}{n} \sum_{i=1}^{n} \frac{|y_i - \hat{y}_i|}{|y_i| + |\hat{y}_i|}
\end{equation}

\noindent where \( y_i \) represents the actual values, \( \hat{y}_i \) represents the predicted values, and \( n \) is the number of observations. sMAPE is capped at 100\%, providing a fixed range. 
This makes sMAPE ideal for challenging tasks like ours, as we can penalize responses that fail to produce an answer with a maximum error of 100\%. 
For a justification of sMAPE over other metrics, see \cref{appendix:metric}.

\subsection{Dataset}
\label{sec:dataset}
\paragraph{\real{}.}
We introduce a set of real images with patterns to test amodal counting in naturalistic settings. 
The original images and annotations come from the FSC-147 dataset \citep{ranjan2021learning}, a diverse counting dataset with manual annotations for the number of target objects and all object bounding boxes in each image. 
FSC-147 contains a diverse array of objects, with 6146 real-world images across 147 object categories. 
We filter FSC-147 for images that contain identifiable and regular patterns of objects and manually overlay a black box to occlude some objects, resulting in 924 images.
Filtering is first performed with GPT-4o and then manually verified; we also manually verify that determining objects despite the occlusion is feasible. 
For each example, we maintain both occluded and unoccluded versions. 
Further details on \real{} can be found in \cref{appendix:dataset}.

\myparagraph{\synthetic{}.} 
While \real{} makes \method{} more applicable to real-world scenarios, each image is unique, making the data less controlled and challenging to draw clear conclusions about model performance. 
Images without background distractors, texture variance, and other potential visual obstacles provide a more controlled version of the task. 
Therefore, we create \synthetic{} to examine the task in a fully controlled environment. 
\synthetic{} comprises 1250 images of simple objects in patterns, where different variables are held constant or changed. 
We vary the following features: 
\begin{enumerate}[noitemsep,topsep=0pt]
    \item \textbf{Object count}: varies from 5 to 15.
    \item \textbf{Object}: can be either dots or squares.
    \item \textbf{Arrangement/shape}: can be a rectangle, circle, or pyramid (where feasible based on object count).
    \item \textbf{Location}: we consider five positions on the page: center, top-left, top-right, bottom-left, or bottom-right.
    \item \textbf{Color}: we randomly choose one of 5 colors for all the objects in an image. 
\end{enumerate}
The \synthetic{} data is split similarly to the \real{} data; each configuration has a variant with an overlaid occluding box and one without.

\subsection{Statistics and Examples}
\begin{figure}[t]
    \centering
    \includegraphics[width=.8\linewidth]{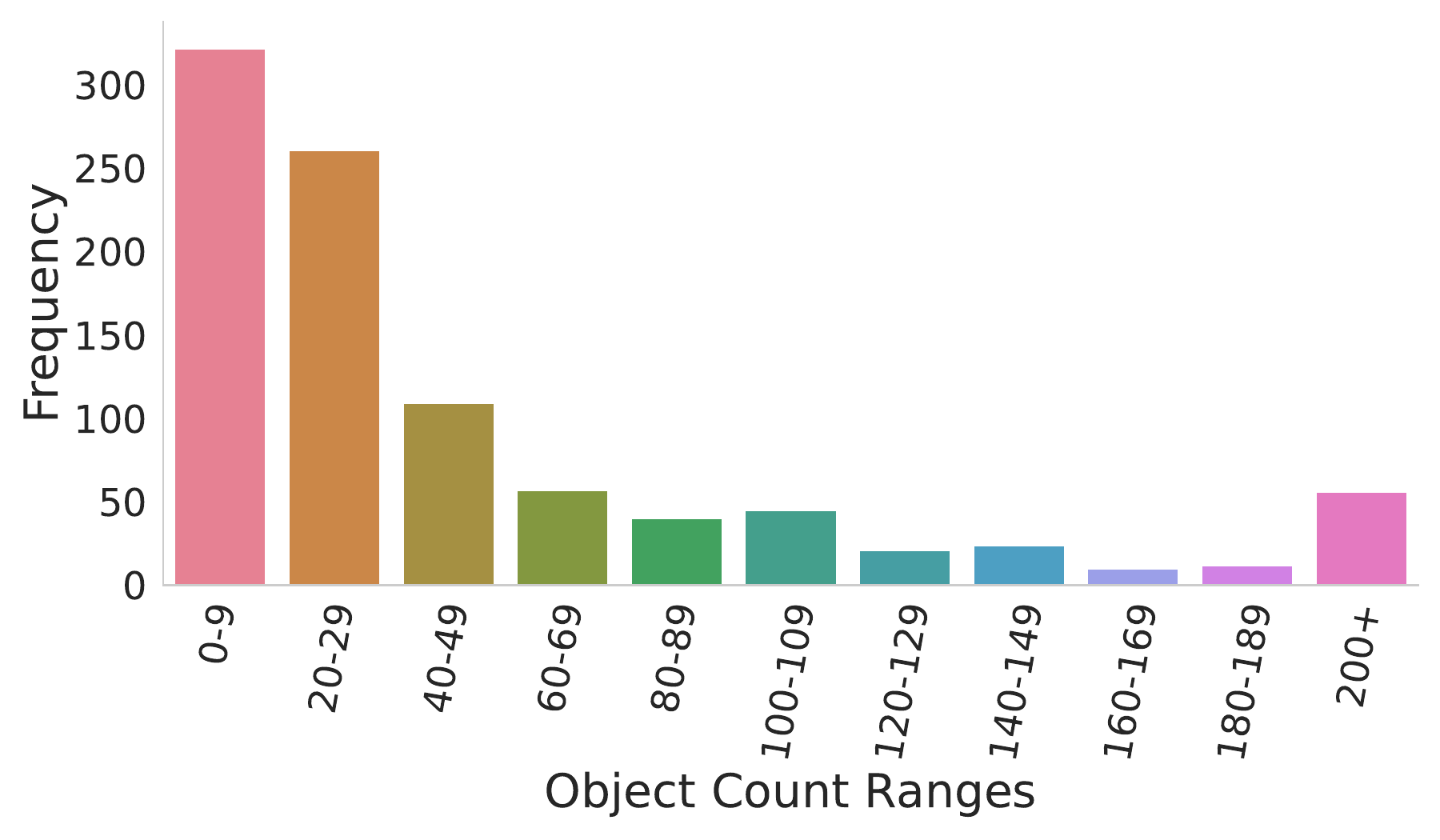}
    \vspace{-0.5em}
    \caption{\# of objects in \real{} images.
    }
    \vspace{-0.5em}
    \label{fig:histogram}
\end{figure}

\begin{figure}[t]
\centering
        \includegraphics[width=.8\linewidth]{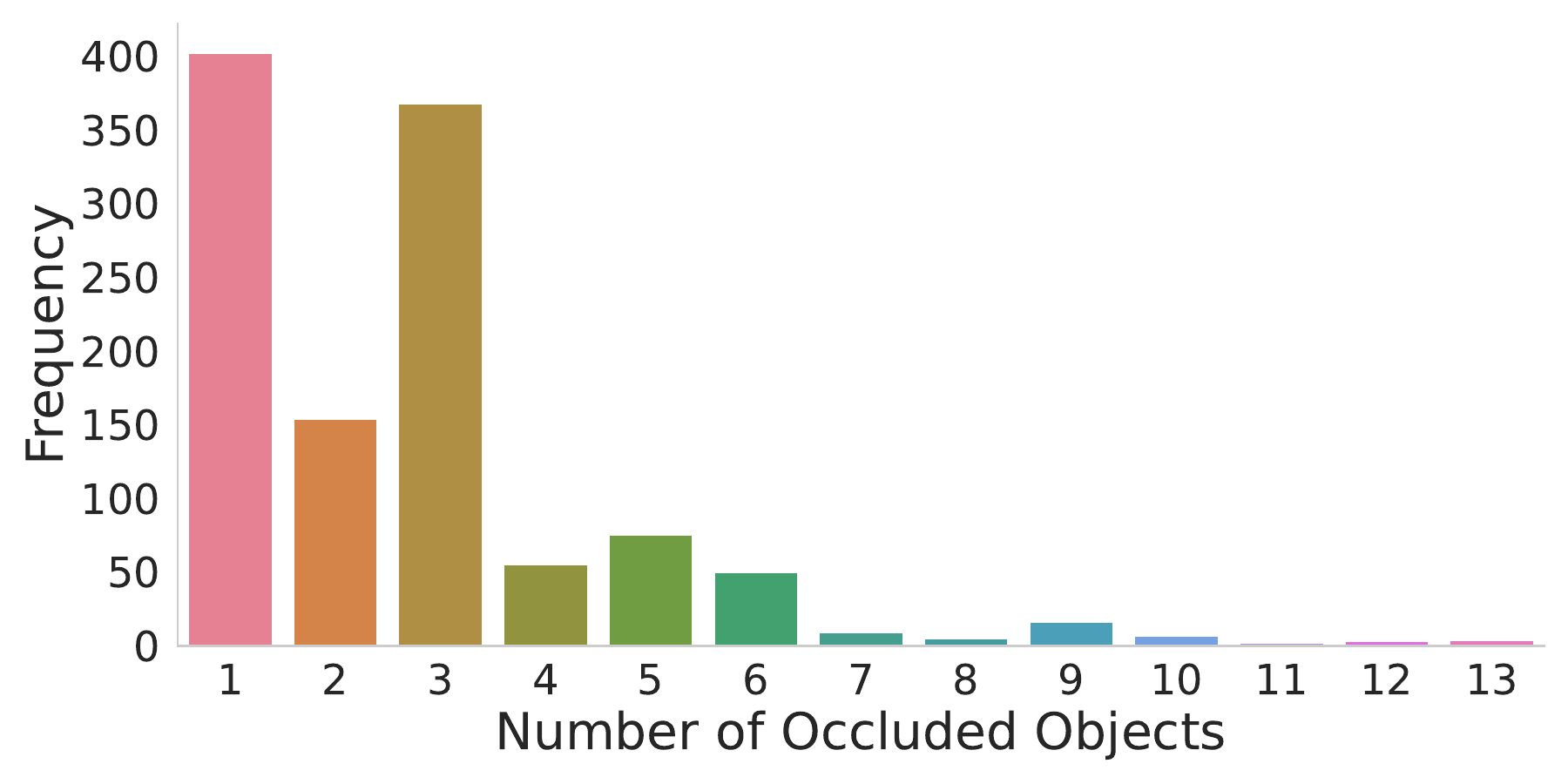}
        \caption{\# of occluded objects in \synthetic{} images.
        }
        \vspace{-1em}
        \label{fig:occ_histogram}
\end{figure}

\cref{fig:examples} shows examples from \real{} and \synthetic{} paired with their corresponding answers from GPT-4o and their ground truth answers. 
These examples show the range of objects and patterns in the dataset and highlight the task's feasibility for humans. 
\cref{tab:dataset} reports summary statistics for \method{}, including the number of images and object types, as well as the mean number of occluded and total objects in both splits of \method{}. 
The number of objects in \real{} is shown in \cref{fig:histogram}, where most images have between 0 and 30 objects. 
On \synthetic{}, the maximum number of objects is 15, and \synthetic{} images generally have 1-6 occluded objects (shown in \cref{fig:occ_histogram}, as further occlusion could make the count unresolvable). 

\begin{table}[t]
\centering
\resizebox{\columnwidth}{!}{
\begin{tabular}{l cc}
\toprule
 & \real{} & \synthetic{} \\
\midrule
\# Images & 924 & 1250\\
\# Object Types & 92 & 2\\
Avg. Occluded Obj. & 13.97 & 2.73\\
Avg. Total Obj. & 61.45 & 10.00\\
\midrule
 & Diverse Objects/Settings & Confounder-free \\
Strengths & Naturalistic & Controllable Attributes \\
 & Realistic Context & Uniformly Distributed\\
\bottomrule
\end{tabular}
}
\caption{Statistics and strengths for \method{} splits.
}
\vspace{-1em}
\label{tab:dataset}
\end{table}

\section{Experiment Setup}
\label{sec:implementation}

\subsection{Models}
We experiment with GPT-4o \citep{gpt4o}, Intern-VL2-Llama3-8B \citep{chen2023internvl, chen2024far}, Qwen2-VL-7B \citep{Qwen2VL}, MiniCPM-o 2.6 \citep{yao2024minicpm}, and Kimi-VL-A3B \citep{kimiteam2025kimivltechnicalreport}
for their high scores on other VLM tasks \citep{openvlm_leaderboard}.
We add Molmo 7B-D \cite{deitke2024molmo}, because of its ability to ``point and count,'' giving it a potential advantage on \method{}.
Specifically, Molmo is trained on millions of examples that directly ground text to 2D coordinates (or ``points'') in images. 
This allows Molmo to directly point to image coordinates and count more easily by pointing to several objects. 
All the VLMs feature a different language backbone and vision encoder to provide broad coverage of model architectures. 
To evaluate models,
we provide the model with the name of the specific object to be counted and the explicit instruction to count fully visible objects and objects behind the occluding box (in the occluded images). 
For each model, we test ten prompts on a validation set of 100 images, selecting the best prompt for each model in each dataset section (\real{}/\synthetic{}) and for each environment (occluded/unoccluded). 
We provide the selected prompts in \cref{appendix:prompts}.

\subsection{Answer Generation and Extraction}
Given the complex nature of \method{}, we allow models to generate open-ended responses and then subsequently extract answers.
Further details (including the maximum number of tokens) can be found in \cref{appendix:output_token}.

\myparagraph{Answer extraction.}
Empirically, we found that constraining the output to a specific format for ease of analysis negatively impacted benchmark performance. 
Therefore, we instead prompt models to generate freely and extract the final output number using a separate answer extractor based on Llama 3.1 8B \citep{llama3modelcard}.
This answer extractor takes the output from the model as input and prompts it to extract a single number representing the final answer. 
The answer extractor also identifies if an output failed to converge on a singular number answer and assigns a label to these examples. 
We mark such incomplete/incoherent model generations as \textit{`skipped'} questions and when calculating the error later, these responses are assigned the worst possible sMAPE score (100\%). 
The answer extractor outputs were manually verified on 1000 outputs, and the extractor was found to be 100\% accurate.

\myparagraph{Human and object detection baselines.}
We also report the performance of humans and a recent counting model (\baseline{} \cite{amini2024countgd}) as baselines to establish a point of reference for model performance.
To confirm that humans can perform the \method{} task, we provided 100 randomly selected occluded examples each from the \real{} and \synthetic{} subsets to 3 undergraduate students with no prior knowledge of the task. 

\section{Results and Analysis}
\label{sec:result}

\subsection{Main Results on \real{}}
\label{sec:real_result}

\begin{table*}[h]
\centering
\resizebox{0.6\textwidth}{!}{%
\begin{tabular}{l cccc}
\toprule
\multirow{3}{*}{Model} & \multicolumn{4}{c}{Error (\%) [$\downarrow$]}\\
\cmidrule(lr){2-5}
& \multicolumn{2}{c}{\real{}} & \multicolumn{2}{c}{\synthetic{}}\\
\cmidrule(lr){2-3} \cmidrule(lr){4-5}
& Original & w/ Occlusion ($\Delta$) & Original & w/ Occlusion ($\Delta$)\\
\midrule
GPT-4o & 13.34 & 14.75 (+1.41) & 5.90 & 9.71 (+3.81)\\
InternVL2 & 26.17 & 32.90 (+6.73) & 16.44 & 17.57 (+1.13)\\
Molmo & 25.90 & 32.49 (+6.59) & 8.40 & 17.73 (+9.33)\\
Qwen2VL & 18.96 & 29.33 (+10.37) & 6.63 & 11.74 (+5.11)\\
MiniCPM-o 2.6 & 23.84 & 30.08 (+6.24) & 17.06 & 19.00 (+1.94)\\
Kimi-VL-A3B & 23.48 & 25.96 (+2.48) & 16.91 & 18.07 (+1.16)\\
\midrule
Avg. of 6 VLMs & 21.95 & 27.59 (+5.64) & 11.89 & 15.64 (+3.75)\\
\bottomrule
\end{tabular}
}
\caption{
Results across VLMs on all splits of \method{}, with average error for each column. Metric: sMAPE (lower is better).
}
\label{tab:real_unocc_occ}
\end{table*}

\begin{table}[h]
    \centering
        \centering
        \resizebox{.9\columnwidth}{!}{%
        \begin{tabular}{l cc}
            \toprule
            \multirow{2}{*}{Model} & \multicolumn{2}{c}{Error (\%) [$\downarrow$]}\\
            \cmidrule(lr){2-3}
             & \real{} & \synthetic{}\\
            \midrule
            \textcolor{gray}{\textit{(Baseline)}}\\
            \textcolor{gray}{Human} & \textcolor{gray}{3.79} & \textcolor{gray}{0.92}\\
            \midrule
            \textcolor{black}{\textit{(VLMs)}}\\
            GPT-4o & 14.75 & 9.71\\
            InternVL2 & 32.90 & 17.57\\
            Molmo & 32.49 & 17.73\\
            Qwen2VL & 29.33 & 11.74\\
            \midrule
            Avg. of 4 VLMs & 27.37 & 14.19\\
            \bottomrule
        \end{tabular}
        }
        \caption{Human baseline vs VLMs on \real{} and \synthetic{} (occluded split). Metric: sMAPE (lower is better).}
        \vspace{-1em}
        \label{tab:human}
\end{table}

\vspace{-0.5em}
\myparagraph{Models consistently struggle with counting and perform worse on occluded images.} 
We run the VLMs on the occluded and unoccluded versions of \method{} to discern whether occlusion significantly impacts model performance. 
\cref{tab:real_unocc_occ} shows that all models struggle with counting generally, performing poorly on both splits. 
Moreover, we see that \textbf{every model performs better on the unoccluded images.} 
On average, the models perform 6.28\% worse in \real{} occluded images and 4.85\% worse in \synthetic{} occluded images (in terms of absolute sMAPE), indicating increased difficulty from a standard counting task. 
The best model for both splits, GPT-4o, has an error rate of 14.75\% on \real{} and a lower error rate of 9.71\% on \synthetic{}.
Across both the real and synthetic split, GPT-4o's error increases with occlusion, by 1.41\% on the real data and 3.81\% on the synthetic split. 
Interestingly, despite its fine-tuning on counting tasks, Molmo exhibits a sizable error rate of 32.5\% on \real{} occluded images.
The high error rates of VLMs indicate limited capabilities in visual understanding under occlusions, pattern recognition, and counting. 
We further analyze the source of these errors with oracle experiments in \cref{sec:oracle_result}. 

\myparagraph{Humans complete the task with almost no error.}
\cref{tab:human}, evaluated on a 100-example subset of each split, confirms that humans complete the task with ease despite occlusion, with an sMAPE of 3.79\% on \real{} and 0.92\% on \synthetic{}.
On the same subset of examples, \textbf{models performed 7 times worse on \real{} and 14 times worse on \synthetic{} than humans}, underscoring the gap between VLMs and humans in this task. 

\myparagraph{Object detection-based baseline outperforms VLMs.}
We attempt the task with a strong object detection-based model to highlight that a standard counting approach will experience a greater loss going from unoccluded to occluded environments, as it cannot capture any occluded objects, i.e. cannot reason. 
We choose \baseline{} \citep{amini2024countgd}, the top solution for unoccluded counting on FSC-147, on which it was trained. 
Because we draw our images from FSC-147's train and test sets, and \baseline{} trains on FSC-147, we only evaluate \baseline{} on the subset of our data sourced from the FSC-147 test set, consisting of 149 images.

\begin{figure}[t]
    \centering
    \includegraphics[width=\columnwidth]{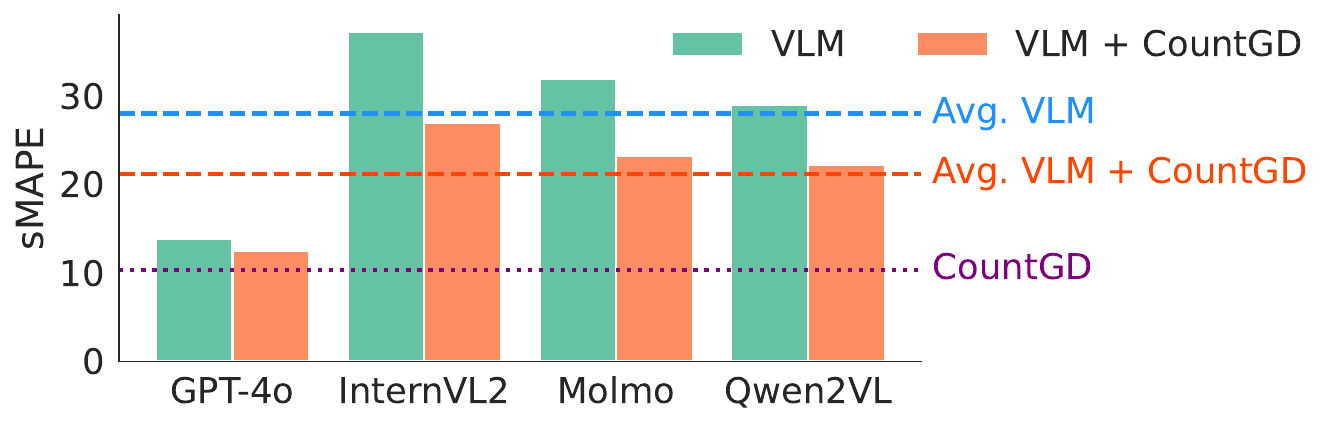}
    \caption{VLM vs. VLM + CountGD hybrid on questions from the \real{} (occluded split) that are not in \baseline{} training set. Metric: sMAPE (lower is better).
    }
    \vspace{-0.5em}
    \label{fig:vlm_countgd}
\end{figure}

We find that \textbf{\baseline{} deteriorates by 7.19\% on occluded images}, increasing from 3.15\% sMAPE to 10.34\% as observed in \cref{fig:vlm_countgd}. 
As expected, \baseline{} outperforms all VLMs on the unoccluded split as it is trained for counting on FSC-147.
\baseline{} also outperforms the VLMs on the occluded split, reinforcing that only counting the visible objects is a hard-to-beat baseline. 
However, the drop in performance with occlusion is greater than the average VLM's drop, highlighting a disadvantage of non-reasoning solutions on \method{}: their error is necessarily tied directly to the number of occluded objects and they cannot address the task on their own, whereas a VLM might be able to infer missing objects via reasoning. 

\begin{figure*}[t]
    \centering
    \includegraphics[width=0.8\linewidth]{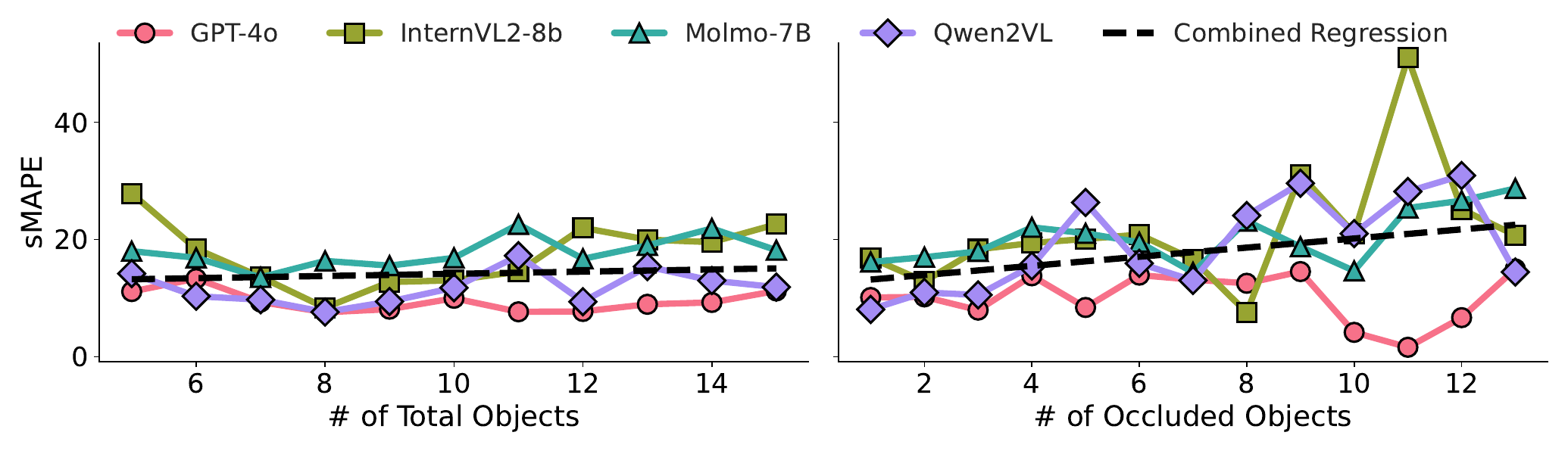}
    \vspace{-0.5em}
    \caption{Effect of number of total objects in the image and number of occluded objects on sMAPE from \synthetic{} (occluded split). Metric: sMAPE (lower is better).
    }
    \vspace{-0.5em}
    \label{fig:num_dots}
\end{figure*}

\myparagraph{Hybrid VLM counting systems improve performance.}
Finding that \baseline{} is far better at counting visible objects than VLMs, we leverage the advantage that \baseline{} has by feeding its visible object count information to the VLMs as part of the prompt. 
As expected, \cref{fig:vlm_countgd} illustrates that there is a considerable decrease in error when CountGD and the VLMs are combined.
However, this hybrid system still performs worse than \baseline{} alone, indicating VLMs are still subpar even at counting just occluded objects (as further reinforced by \cref{appendix:only_occ}).

\subsection{Effect of Data Factors on VLM Performance}
Here, we use the \synthetic{} data (which can be controlled precisely and 
fully annotated
) to examine which features correlate with model performance. 
We test the effect of the following variables on final performance:
(1) Increasing the number of occluded objects; (2) Varying the pattern. We also investigate 
whether models can classify patterns, and to what degree models can predict the number of occluded objects only (rather than the total). 

\myparagraph{Models perform worse when more dots are occluded.} 
In \cref{fig:num_dots} (right), we observe that \textbf{error increases with respect to the number of occluded dots.} 
However, \cref{fig:num_dots} (left) also shows that performance is less affected by the total number of dots.
This suggests that the task difficulty is more closely correlated with the difficulty of occlusion -- i.e. the difficulty of the world modeling task -- rather than the complexity of the pattern. 
Some models, such as GPT-4o, deviate from this trend, which has lower error on specific numbers. 
We further explore model bias in \cref{appendix:heatmap}. 

\begin{figure}[t]
    \centering
    \includegraphics[width=0.9\linewidth]{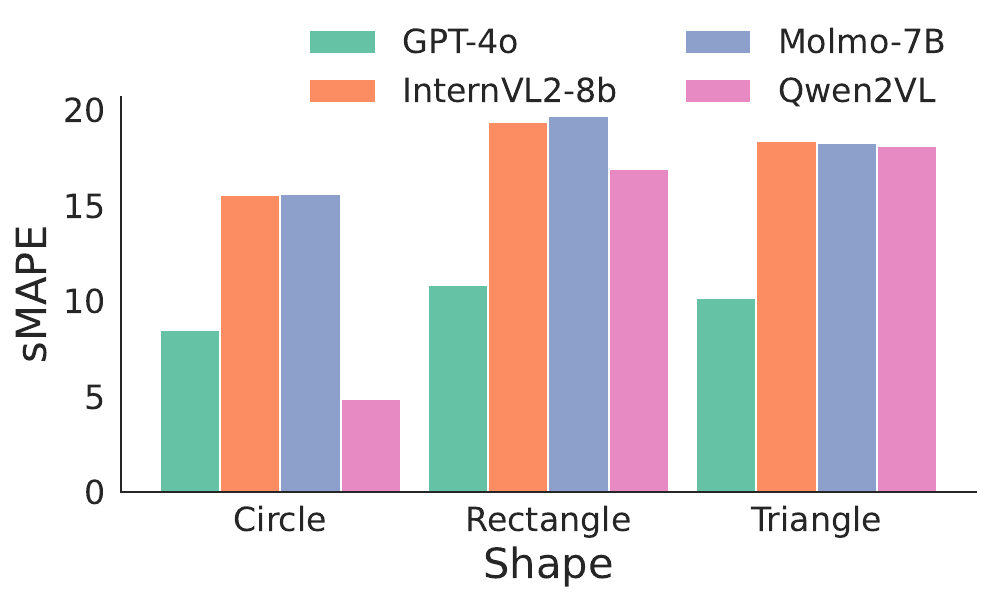}
    \vspace{-0.5em}
    \caption{Effect of pattern type in \synthetic{} (occluded split) on sMAPE. Metric: sMAPE (lower is better). 
    }
    \vspace{-1em}
    \label{fig:shape_graph}
\end{figure}

\myparagraph{Performance depends on pattern type.}
The controllability of \synthetic{} allows us to measure the effect of pattern type on performance.
In \cref{fig:shape_graph}, we find that model performance differs across shapes with some regularity: objects arranged in a circle generally have lower sMAPE than other shapes, across all models.  
\begin{table}[h]
\centering
\resizebox{0.8\columnwidth}{!}{%
\begin{tabular}{l cc}
\toprule
\multirow{2}{*}{Model} & \multicolumn{2}{c}{Accuracy (\%) [$\uparrow$]} \\
\cmidrule(lr){2-3}
 & Original & w/ Occlusion ($\Delta$) \\
\midrule
GPT-4o & 84.00 & 78.52 (-5.48) \\
InternVL2 & 68.52 & 47.48 (-21.04) \\
Molmo & 80.70 & 65.22 (-15.48) \\
Qwen2VL & 88.35 & 86.43 (-1.92) \\
\midrule
Avg. of 4 VLMs & 80.39 & 69.41 (-10.98) \\
\bottomrule
\end{tabular}
}
\caption{
VLM accuracy in identifying the correct pattern in \synthetic{}. Metric: accuracy (higher is better).
}
\label{tab:shape_table}
\vspace{-0.5em}
\end{table}
Qwen2VL has an especially large decrease in error when given circular arrangements compared to rectangles or triangles.

\begin{figure*}[h]
    \centering
    \includegraphics[width=\textwidth]{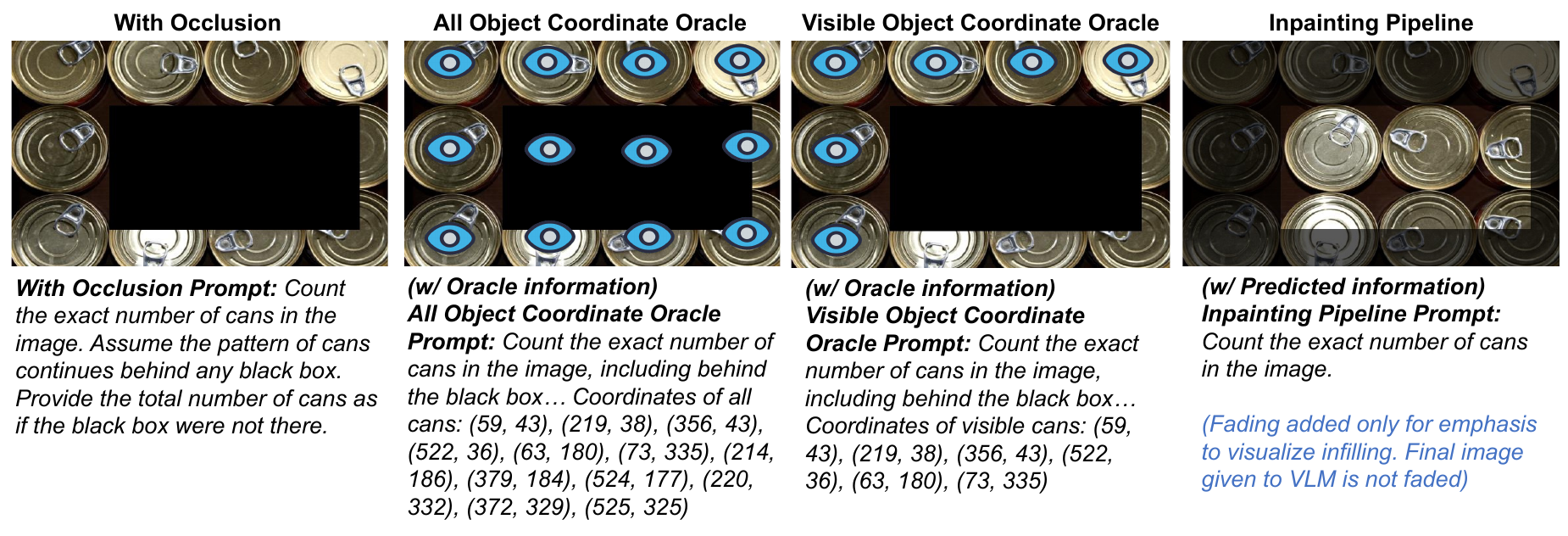}
    \vspace{-20pt}
    \caption{Example image and text inputs for experiments with auxiliary information experiments (\cref{sec:oracle_result}). Blue eyes indicate objects for which the \emph{All Object Coordinate Oracle} or \emph{Visible Object Coordinate Oracle} extracts coordinates. The brighter part of the image represents the area which \emph{\inpaint{}} fills in.
    Example prompts are shown in italics. 
    Blue eye overlays and faded parts of images are for demonstration purposes and are not passed with the image. 
    }
    \label{fig:oracle}
\end{figure*}

\begin{table*}[t]
\centering
\resizebox{.83\textwidth}{!}{%
\begin{tabular}{lc|cccc}
\toprule
\multirow{2}{*}{Model} & \multirow{2}{*}{\textcolor{gray}{Original}} & \multirow{2}{*}{w/ Occlusion} & \multicolumn{2}{c}{Oracle Information} & \multicolumn{1}{c}{Predicted Information} \\
\cmidrule(lrr){4-5} \cmidrule(lrr){6-6}
& & & + All Coordinates ($\Delta$) & + Visible ($\Delta$) & + Inpainting ($\Delta$) \\
\midrule
GPT-4o          & \textcolor{gray}{13.34} & 14.75 & 2.93 (-11.82)  & 9.20 (-5.55)   & 15.89 (+1.14) \\
InternVL2       & \textcolor{gray}{26.17} & 32.90 & 17.48 (-15.42) & 25.13 (-7.77)  & 31.12 (-1.78) \\
Qwen2VL         & \textcolor{gray}{18.96} & 29.33 & 9.62 (-19.71)  & 17.70 (-11.63) & 22.64 (-6.69) \\
\midrule
Avg.\ of 3 VLMs & \textcolor{gray}{19.49} & 25.66 & 10.01 (-15.65) & 17.34 (-8.32)  & 23.22 (-2.44) \\
\bottomrule
\end{tabular}
}
\caption{
Effect of auxiliary information on occluded \real{}. \(\Delta =\) (Auxiliary Information) \(-\) (w/ Occlusion). Metric: sMAPE.
}
\label{tab:oracle}
\end{table*}

\myparagraph{Models can identify patterns.}
To determine how much model errors can be attributed to a lack of pattern recognition ability, we formulate a separate task where models must recognize the pattern in the image on \synthetic{}.
Here, we frame the task as multiple-choice, asking the model to select from the pattern types available (rectangle, triangle, or circle).
\Cref{tab:shape_table} illustrates that all perform substantially better than random at this task, with most models except InternVL2 achieving accuracy above 80\% in the unoccluded setting. 
As expected, the patterns were easier to identify in unoccluded scenarios, with models suffering an average accuracy drop of 10.95\% in the occluded setting. 
Notably, GPT-4o and Qwen2VL have a fairly small drop in performance, suggesting they can generally capture the pattern even in the presence of occlusion.

\subsection{Analysis with Auxiliary Information} 
\label{sec:oracle_result}
In \cref{sec:real_result}, we see that models broadly struggle with amodal counting.
Here, we seek to disentangle whether this problem results from a failure to reason, the absence of a world model, or both by giving VLMs two different types of auxiliary information: \textit{oracle} information and \textit{predicted} information.
Oracle information is ground truth and is directly pulled from \method{}'s metadata, e.g., object locations. 
Predicted information generates new information from a completely separate model and gives it to the VLM. 
This information is not ground truth and is sourced from an external model, such as an image inpainting model, rather than the VLM.
By giving the model auxiliary information in the form of reasoning and spatial clues, we can establish how much of each model's error results from an inability to handle occlusion rather than an inability to recognize and count visible objects. 

\myparagraph{Oracle setup.}
We test two oracles for \real{}'s occluded split based on its constituent subtasks: counting the visible objects and inferring/counting occluded objects. 
Both oracles provide the VLM with text-based coordinates of objects in the image, simplifying the visual task by assuming the VLM effectively has a perfect visual system that can recognize and localize objects in the image. 
The first oracle, the \emph{Visible Object Coordinate Oracle}, gives the VLM the coordinates of all unoccluded objects (encoded as text, as seen in \cref{fig:oracle}) 
and instructs the model to estimate the number of occluded objects, count the number of visible object coordinates, and add the two. 
In other words, the model is given oracle information about what objects are visible, thus also revealing key information about the pattern. 
The second oracle, the \emph{All Object Coordinate Oracle}, instead gives the model the coordinates of all objects.
Here, the model only needs to count the coordinates in the prompt, eliminating the need to reason on the visual input. Note that Molmo is excluded in these tests because it contains a prompt limit that would truncate the list of coordinates.
An example of the oracle inputs can be seen in \cref{fig:oracle}.

\myparagraph{Prediction setup.}
In this setting, we provide the VLM with an external world model representation predicted by another model.
Specifically, we develop the \emph{\inpaint{}} to fill in the occluded region via a diffusion-based inpainting model and pass the inpainted image to the VLMs. 
For the inpainting model, we choose FLUX.1-Fill [dev], whose backbone FLUX.1 [dev] \cite{flux2024} is a top public model in the Text to Image Model Arena \cite{aa_arena2025}. 
An example input to the VLM can be seen on the far-right of \cref{fig:oracle}.

\myparagraph{Providing visible or all object coordinates improves performance substantially.}
The results in \cref{tab:real_unocc_occ} indicate that models struggle on \method{}, which requires identifying a pattern and counting both visible and occluded objects.
Moreover, models generally struggle with counting even in unoccluded settings. 
Both oracles simplify the counting task: \emph{All Object Coordinate Oracle} reduces the task to simply counting coordinates with no reasoning involved, and \emph{Visible Object Coordinate Oracle} similarly simplifies the task for visible objects, while still requiring inferring occluded objects. 
Additionally, under \emph{Visible Object Coordinate Oracle}, recognizing the pattern shifts from a visual reasoning task to an augmented math problem. 
Instead of visually reasoning about where objects are located, the VLM considers what patterns the coordinates could make. 
Translating this task into a text problem results in an average increase of 15\% with all objects coordinate oracle;
the errors LLMs make here are due to an inability to count in the text prompt, as opposed to weaknesses in handling occlusion (since all object coordinates are given), and the strongest model, GPT-4o, achieves minimal error here.
We also obtain an average increase of 8\% with the visible objects coordinate oracle (shown in \cref{tab:oracle}), possibly because it allows the more powerful LLM backbone (which is far larger than the vision model in all models tested) to complete the counting task. 
Taken together, these results suggest that there is much room for improvement in visual world modeling beyond text-based reasoning of VLMs.

\myparagraph{Providing diffusion-based inpainting improves performance marginally.}
Similar to the object coordinate oracles, the \inpaint{} (rightmost columns in \cref{fig:oracle} and \cref{tab:oracle}) eliminates the need for world modeling and provides VLMs with an approximation of the image behind the occluder. 
With the inpainted images, VLM error decreases by almost 2\% for InternVL2 and 7\% for Qwen2VL compared to the original occluded images. 
GPT-4o's error increases on inpainted images by a small margin; we hypothesize that this may be because GPT-4o has one of the better world models (based on its superior performance), and thus does not improve further with the inpainted images. 
Moreover, every VLM still falls short of its unoccluded image performance, indicating that the diffusion model is not a perfect world model. 
Qualitatively, we find that the inpainting model sometimes fails to output the correct pattern.

\section{Related Work}
\label{sec:related}
\vspace{-0.5em}
\myparagraph{Spatial reasoning in visual question answering.}
Past work measures the spatial reasoning capabilities of VLMs in the form of visual question answering (VQA)~\citep{antol2015vqa, goyal2017making} benchmarks.
SpartQA~\cite{mirzaee2021spartqa} asks VLMs to identify the spatial relation (e.g., above, behind, left of) between objects in synthetically created 2D images from NLVR~\cite{suhr-etal-2017-corpus}.
More recent benchmarks test similar spatial relation understanding with real images~\cite{Liu2022VisualSR, rajabi2024gsr, al2024unibench}.
While this past work asks models to provide a text description for a relation between two fully observed objects,
\method{} measures the world modeling from a partially observed scene, thus requiring the handling of occlusion, pattern recognition, and counting. 
Together, these constitute a stricter test of spatial reasoning than typical VQA settings.

\myparagraph{Amodal completion.}
Occlusions are common in natural scenes, and vision solutions for amodal completion have made significant progress in infilling occlusions \citep{xu2024amodal, saleh2024mask, ao2024open}.
The amodal completion task has evolved from simply completing a shape to filling in appearance (e.g., texture, color, etc.) to finally dealing with fine-grained order perception (multiple stacked occluded objects) \citep{ao2023image}. 
Specifically in \citet{qiu2024occ}, VLMs classify the hidden objects and extract fine details from occluded items.
\method{}, however, presents a unique category of \emph{patterned} amodal counting which requires inferring fully occluded objects based on a pattern rather than inferring occluded object wholes based on object parts.
In other words, previous work has only attempted tasks that require amodal completion for one object at a time \citep{xu2024amodal, ozguroglu2024pix2gestalt, saleh2024mask}, whereas \method{} handles multiple objects. 
Multi-object amodal completion is crucial because in cluttered scenes, entire groups of objects are often occluded.
Moreover, the output space of \method{} is language (rather than filling pixels).

\myparagraph{Counting with vision-and-language models.}
Within the task of counting, the most similar application to \method{} is dense counting, where the objects to be counted occlude each other. 
There are many practical applications of such a task, like counting cells on a crowded slide \citep{bera2015partially}, determining crop yields from densely-packed fields \citep{wang2021occlusion}, or crowd counting \citep{wang2024dual, zhou2024multi, fan2023multi}. 
\citet{liang2023crowdclip} improved crowd counting with an augmented CLIP \cite{radford2021learning}, i.e. also using VLMs for counting. 
Additionally, \citet{jenkins2023countnet3d} introduced an amodal counting benchmark, presenting an occluded 3D counting task where models must count objects on retail shelves. However, our work differs in many ways, as \citet{jenkins2023countnet3d} only counts retail shelves and uses LiDAR input. 
More broadly, dense counting focuses on overlapping objects rather than on counting objects arranged into patterns, which is the focus of \method{}.

\section{Conclusion}
\label{sec:conclusion}
We introduced \method{}, a novel benchmark for amodal counting that measures spatial reasoning capabilities under occlusion. 
\method{} is designed to assess VLMs' ability to form a robust world model and use that model for visual reasoning skills under occlusion.
By testing counting, we cast the problem as a measurable task with an objective correct answer that also has real-world utility as VLMs become more broadly adopted.
Our results suggest that \textbf{VLMs struggle to combine reasoning, counting, and world modeling} with low performance on occluded and unoccluded images. 
Our analysis indicates that models improve with oracle information about visible objects (simplifying the reasoning/counting tasks) and predicted information about the occluded objects (also simplifying world modeling), pointing to directions of model improvement.

\section*{Acknowledgments}
\label{sec:acknowledge}

This work was supported by DARPA ECOLE Program No. HR00112390060,
NSF-CAREER Award 1846185, NSF-AI Engage Institute DRL-2112635, DARPA Machine Commonsense (MCS) Grant N66001-19-2-4031, ARO Award W911NF2110220, ONR Grant N00014-23-1-2356, Microsoft Accelerate Foundation Models Research (AFMR) grant program, and a Bloomberg Data Science PhD Fellowship. The views contained in this article are those of the authors and not of the funding agency.

{
    \small
    \bibliographystyle{ieeenat_fullname}
    \bibliography{main}

\begin{thebibliography}{48}
\providecommand{\natexlab}[1]{#1}
\providecommand{\url}[1]{\texttt{#1}}
\expandafter\ifx\csname urlstyle\endcsname\relax
  \providecommand{\doi}[1]{doi: #1}\else
  \providecommand{\doi}{doi: \begingroup \urlstyle{rm}\Url}\fi

\bibitem[AI@Meta(2024)]{llama3modelcard}
AI@Meta.
\newblock Llama 3.1 model card.
\newblock \emph{Github Model Card}, 2024.

\bibitem[Al-Tahan et~al.(2024)Al-Tahan, Garrido, Balestriero, Bouchacourt, Hazirbas, and Ibrahim]{al2024unibench}
Haider Al-Tahan, Quentin Garrido, Randall Balestriero, Diane Bouchacourt, Caner Hazirbas, and Mark Ibrahim.
\newblock Unibench: Visual reasoning requires rethinking vision-language beyond scaling.
\newblock \emph{arXiv preprint arXiv:2408.04810}, 2024.

\bibitem[Amini-Naieni et~al.(2024)Amini-Naieni, Han, and Zisserman]{amini2024countgd}
Niki Amini-Naieni, Tengda Han, and Andrew Zisserman.
\newblock Countgd: Multi-modal open-world counting.
\newblock \emph{arXiv preprint arXiv:2407.04619}, 2024.

\bibitem[Antol et~al.(2015)Antol, Agrawal, Lu, Mitchell, Batra, Zitnick, and Parikh]{antol2015vqa}
Stanislaw Antol, Aishwarya Agrawal, Jiasen Lu, Margaret Mitchell, Dhruv Batra, C~Lawrence Zitnick, and Devi Parikh.
\newblock Vqa: Visual question answering.
\newblock In \emph{Proceedings of the IEEE international conference on computer vision}, pages 2425--2433, 2015.

\bibitem[Ao et~al.(2023)Ao, Ke, and Ehinger]{ao2023image}
Jiayang Ao, Qiuhong Ke, and Krista~A Ehinger.
\newblock Image amodal completion: A survey.
\newblock \emph{Computer Vision and Image Understanding}, 229:\penalty0 103661, 2023.

\bibitem[Ao et~al.(2024)Ao, Jiang, Ke, and Ehinger]{ao2024open}
Jiayang Ao, Yanbei Jiang, Qiuhong Ke, and Krista~A Ehinger.
\newblock Open-world amodal appearance completion.
\newblock \emph{arXiv preprint arXiv:2411.13019}, 2024.

\bibitem[{Artificial Analysis}(2025)]{aa_arena2025}
{Artificial Analysis}.
\newblock Text to image model arena, 2025.
\newblock Accessed: April 10, 2025.

\bibitem[Bera(2015)]{bera2015partially}
Soumen Bera.
\newblock Partially occluded object detection and counting.
\newblock In \emph{Proceedings of the 2015 Third International Conference on Computer, Communication, Control and Information Technology (C3IT)}, pages 1--6. IEEE, 2015.

\bibitem[Chen et~al.(2023)Chen, Wu, Wang, Su, Chen, Xing, Zhong, Zhang, Zhu, Lu, Li, Luo, Lu, Qiao, and Dai]{chen2023internvl}
Zhe Chen, Jiannan Wu, Wenhai Wang, Weijie Su, Guo Chen, Sen Xing, Muyan Zhong, Qinglong Zhang, Xizhou Zhu, Lewei Lu, Bin Li, Ping Luo, Tong Lu, Yu Qiao, and Jifeng Dai.
\newblock Internvl: Scaling up vision foundation models and aligning for generic visual-linguistic tasks.
\newblock \emph{arXiv preprint arXiv:2312.14238}, 2023.

\bibitem[Chen et~al.(2024)Chen, Wang, Tian, Ye, Gao, Cui, Tong, Hu, Luo, Ma, et~al.]{chen2024far}
Zhe Chen, Weiyun Wang, Hao Tian, Shenglong Ye, Zhangwei Gao, Erfei Cui, Wenwen Tong, Kongzhi Hu, Jiapeng Luo, Zheng Ma, et~al.
\newblock How far are we to gpt-4v? closing the gap to commercial multimodal models with open-source suites.
\newblock \emph{arXiv preprint arXiv:2404.16821}, 2024.

\bibitem[Chicco et~al.(2021)Chicco, Warrens, and Jurman]{chicco2021coefficient}
Davide Chicco, Matthijs~J Warrens, and Giuseppe Jurman.
\newblock The coefficient of determination r-squared is more informative than smape, mae, mape, mse and rmse in regression analysis evaluation.
\newblock \emph{Peerj computer science}, 7:\penalty0 e623, 2021.

\bibitem[Coupland(2011)]{coupland2011frequent}
Nikolas Coupland.
\newblock How frequent are numbers?
\newblock \emph{Language \& Communication}, 31\penalty0 (1):\penalty0 27--37, 2011.

\bibitem[Deitke et~al.(2024)Deitke, Clark, Lee, Tripathi, Yang, Park, Salehi, Muennighoff, Lo, Soldaini, et~al.]{deitke2024molmo}
Matt Deitke, Christopher Clark, Sangho Lee, Rohun Tripathi, Yue Yang, Jae~Sung Park, Mohammadreza Salehi, Niklas Muennighoff, Kyle Lo, Luca Soldaini, et~al.
\newblock Molmo and pixmo: Open weights and open data for state-of-the-art multimodal models.
\newblock \emph{arXiv preprint arXiv:2409.17146}, 2024.

\bibitem[Fan et~al.(2023)Fan, Song, Wu, and Zhu]{fan2023multi}
Zheyi Fan, Zihao Song, Di Wu, and Yixuan Zhu.
\newblock Multi-branch segmentation-guided attention network for crowd counting.
\newblock \emph{Journal of Visual Communication and Image Representation}, 97:\penalty0 103964, 2023.

\bibitem[Flores(1986)]{flores1986pragmatic}
Benito~E Flores.
\newblock A pragmatic view of accuracy measurement in forecasting.
\newblock \emph{Omega}, 14\penalty0 (2):\penalty0 93--98, 1986.

\bibitem[Goyal et~al.(2017)Goyal, Khot, Summers-Stay, Batra, and Parikh]{goyal2017making}
Yash Goyal, Tejas Khot, Douglas Summers-Stay, Dhruv Batra, and Devi Parikh.
\newblock Making the v in vqa matter: Elevating the role of image understanding in visual question answering.
\newblock In \emph{Proceedings of the IEEE conference on computer vision and pattern recognition}, pages 6904--6913, 2017.

\bibitem[Ha and Schmidhuber(2018)]{ha2018recurrent}
David Ha and J{\"u}rgen Schmidhuber.
\newblock Recurrent world models facilitate policy evolution.
\newblock \emph{Advances in neural information processing systems}, 31, 2018.

\bibitem[Jenkins et~al.(2023)Jenkins, Armstrong, Nelson, Gotad, Jenkins, Wilkey, and Watts]{jenkins2023countnet3d}
Porter Jenkins, Kyle Armstrong, Stephen Nelson, Siddhesh Gotad, J~Stockton Jenkins, Wade Wilkey, and Tanner Watts.
\newblock Countnet3d: A 3d computer vision approach to infer counts of occluded objects.
\newblock In \emph{Proceedings of the IEEE/CVF Winter Conference on Applications of Computer Vision}, pages 3008--3017, 2023.

\bibitem[Kanizsa et~al.(1979)Kanizsa, Legrenzi, and Bozzi]{gestalt_perception}
Gaetano Kanizsa, Paolo Legrenzi, and Paolo Bozzi.
\newblock \emph{Organization in vision : essays on gestalt perception}.
\newblock Praeger, 1979.

\bibitem[Kassaw et~al.(2024)Kassaw, Luzi, Collins, and Malof]{kassaw2024deep}
Kaleb Kassaw, Francesco Luzi, Leslie~M Collins, and Jordan~M Malof.
\newblock Are deep learning models robust to partial object occlusion in visual recognition tasks?
\newblock \emph{arXiv preprint arXiv:2409.10775}, 2024.

\bibitem[Labs(2024)]{flux2024}
Black~Forest Labs.
\newblock Flux.
\newblock \url{https://github.com/black-forest-labs/flux}, 2024.

\bibitem[Li et~al.(2024)Li, Lin, Peng, Nyandwi, Jiang, Ma, Khanuja, Krishna, Neubig, and Ramanan]{li2024naturalbench}
Baiqi Li, Zhiqiu Lin, Wenxuan Peng, Jean de~Dieu Nyandwi, Daniel Jiang, Zixian Ma, Simran Khanuja, Ranjay Krishna, Graham Neubig, and Deva Ramanan.
\newblock Naturalbench: Evaluating vision-language models on natural adversarial samples.
\newblock \emph{arXiv preprint arXiv:2410.14669}, 2024.

\bibitem[Liang et~al.(2023)Liang, Xie, Zou, Ye, Xu, and Bai]{liang2023crowdclip}
Dingkang Liang, Jiahao Xie, Zhikang Zou, Xiaoqing Ye, Wei Xu, and Xiang Bai.
\newblock Crowdclip: Unsupervised crowd counting via vision-language model.
\newblock In \emph{Proceedings of the IEEE/CVF conference on computer vision and pattern recognition}, pages 2893--2903, 2023.

\bibitem[Liu et~al.(2023)Liu, Emerson, and Collier]{Liu2022VisualSR}
Fangyu Liu, Guy Edward~Toh Emerson, and Nigel Collier.
\newblock Visual spatial reasoning.
\newblock \emph{Transactions of the Association for Computational Linguistics}, 2023.

\bibitem[Maiseli(2019)]{maiseli2019optimum}
Baraka~Jacob Maiseli.
\newblock Optimum design of chamfer masks using symmetric mean absolute percentage error.
\newblock \emph{EURASIP Journal on Image and Video Processing}, 2019\penalty0 (1):\penalty0 74, 2019.

\bibitem[Mirzaee and Rajaby(2021)]{mirzaee2021spartqa}
Roshanak Mirzaee and Hossein Rajaby.
\newblock Spartqa: A textual question answering benchmark for spatial reasoning.
\newblock In \emph{The 2021 Annual Conference of the North American Chapter of the Association for Computational Linguistics (NAACL-2021)}, 2021.

\bibitem[Olson et~al.(2004)Olson, Gatenby, Leung, Skudlarski, and Gore]{olson2004neuronal}
Ingrid~R Olson, J~Christopher Gatenby, Hoi-Chung Leung, Pawel Skudlarski, and John~C Gore.
\newblock Neuronal representation of occluded objects in the human brain.
\newblock \emph{Neuropsychologia}, 42\penalty0 (1):\penalty0 95--104, 2004.

\bibitem[{OpenAI}(2024)]{gpt4o}
{OpenAI}.
\newblock Hello gpt-4o, 2024.

\bibitem[{OpenCompass Team}(2024)]{openvlm_leaderboard}
{OpenCompass Team}.
\newblock Openvlm leaderboard.
\newblock \url{https://huggingface.co/spaces/opencompass/open_vlm_leaderboard}, 2024.
\newblock Accessed: 2024-11-13.

\bibitem[OTSUKA et~al.(2006)OTSUKA, KANAZAWA, and YAMAGUCHI]{otsuka2006development}
Yumiko OTSUKA, So KANAZAWA, and Masami~K YAMAGUCHI.
\newblock Development of modal and amodal completion in infants.
\newblock \emph{Perception (London. Print)}, 35\penalty0 (9):\penalty0 1251--1264, 2006.

\bibitem[Ozguroglu et~al.(2024)Ozguroglu, Liu, Sur{\'\i}s, Chen, Dave, Tokmakov, and Vondrick]{ozguroglu2024pix2gestalt}
Ege Ozguroglu, Ruoshi Liu, D{\'\i}dac Sur{\'\i}s, Dian Chen, Achal Dave, Pavel Tokmakov, and Carl Vondrick.
\newblock pix2gestalt: Amodal segmentation by synthesizing wholes.
\newblock In \emph{2024 IEEE/CVF Conference on Computer Vision and Pattern Recognition (CVPR)}, pages 3931--3940. IEEE Computer Society, 2024.

\bibitem[Peeperkorn et~al.(2024)Peeperkorn, Kouwenhoven, Brown, and Jordanous]{peeperkorn2024temperature}
Max Peeperkorn, Tom Kouwenhoven, Dan Brown, and Anna Jordanous.
\newblock Is temperature the creativity parameter of large language models?
\newblock \emph{arXiv preprint arXiv:2405.00492}, 2024.

\bibitem[Qharabagh et~al.(2024)Qharabagh, Ghofrani, and Fountoulakis]{qharabagh2024lvlm}
Muhammad~Fetrat Qharabagh, Mohammadreza Ghofrani, and Kimon Fountoulakis.
\newblock Lvlm-count: Enhancing the counting ability of large vision-language models.
\newblock \emph{arXiv preprint arXiv:2412.00686}, 2024.

\bibitem[Qiu and Di(2024)]{qiu2024occ}
Wenmo Qiu and Xinhan Di.
\newblock Occ-mllm: Empowering multimodal large language model for the understanding of occluded objects.
\newblock \emph{arXiv preprint arXiv:2410.01261}, 2024.

\bibitem[Radford et~al.(2021)Radford, Kim, Hallacy, Ramesh, Goh, Agarwal, Sastry, Askell, Mishkin, Clark, et~al.]{radford2021learning}
Alec Radford, Jong~Wook Kim, Chris Hallacy, Aditya Ramesh, Gabriel Goh, Sandhini Agarwal, Girish Sastry, Amanda Askell, Pamela Mishkin, Jack Clark, et~al.
\newblock Learning transferable visual models from natural language supervision.
\newblock In \emph{International conference on machine learning}, pages 8748--8763. PmLR, 2021.

\bibitem[Rajabi and Kosecka(2024)]{rajabi2024gsr}
Navid Rajabi and Jana Kosecka.
\newblock Gsr-bench: A benchmark for grounded spatial reasoning evaluation via multimodal llms.
\newblock \emph{arXiv preprint arXiv:2406.13246}, 2024.

\bibitem[Ranjan et~al.(2021)Ranjan, Sharma, Nguyen, and Hoai]{ranjan2021learning}
Viresh Ranjan, Udbhav Sharma, Thu Nguyen, and Minh Hoai.
\newblock Learning to count everything.
\newblock In \emph{Proceedings of the IEEE/CVF Conference on Computer Vision and Pattern Recognition}, pages 3394--3403, 2021.

\bibitem[Saleh et~al.(2024)Saleh, Sz{\'e}n{\'a}si, and V{\'a}mossy]{saleh2024mask}
Kaziwa Saleh, S{\'a}ndor Sz{\'e}n{\'a}si, and Zolt{\'a}n V{\'a}mossy.
\newblock Mask guided gated convolution for amodal content completion.
\newblock In \emph{2024 IEEE 22nd Jubilee International Symposium on Intelligent Systems and Informatics (SISY)}, pages 000321--000326. IEEE, 2024.

\bibitem[Suhr et~al.(2017)Suhr, Lewis, Yeh, and Artzi]{suhr-etal-2017-corpus}
Alane Suhr, Mike Lewis, James Yeh, and Yoav Artzi.
\newblock A corpus of natural language for visual reasoning.
\newblock In \emph{Proceedings of the 55th Annual Meeting of the Association for Computational Linguistics (Volume 2: Short Papers)}, pages 217--223, Vancouver, Canada, 2017. Association for Computational Linguistics.

\bibitem[Team et~al.(2025)Team, Du, Yin, Xing, Qu, Wang, Chen, Zhang, Du, Wei, Wang, Zhang, Du, Wang, Yuan, Lu, Li, Sung, Wei, Lai, Zhu, Ding, Hu, Yang, Zhang, Wu, Yao, Lu, Wang, Gao, Zheng, Li, Su, Wang, Deng, Qiu, Xie, Wang, Liu, Yan, Ouyang, Chen, Sui, Yu, Dong, Dong, Xu, Cheng, Gu, Zhou, Liu, Cao, Yu, Song, Bai, Song, He, Huang, Xu, Yuan, Yao, Wu, Zu, Zhou, Wang, Charles, Zhong, Li, Hu, Chen, Wang, Liu, Miao, Qin, Chen, Bao, Wang, Kang, Liu, Du, Wu, Wang, Yan, Zhou, Li, Jiang, Zhang, Yang, Huang, Huang, Zhao, and Chen]{kimiteam2025kimivltechnicalreport}
Kimi Team, Angang Du, Bohong Yin, Bowei Xing, Bowen Qu, Bowen Wang, Cheng Chen, Chenlin Zhang, Chenzhuang Du, Chu Wei, Congcong Wang, Dehao Zhang, Dikang Du, Dongliang Wang, Enming Yuan, Enzhe Lu, Fang Li, Flood Sung, Guangda Wei, Guokun Lai, Han Zhu, Hao Ding, Hao Hu, Hao Yang, Hao Zhang, Haoning Wu, Haotian Yao, Haoyu Lu, Heng Wang, Hongcheng Gao, Huabin Zheng, Jiaming Li, Jianlin Su, Jianzhou Wang, Jiaqi Deng, Jiezhong Qiu, Jin Xie, Jinhong Wang, Jingyuan Liu, Junjie Yan, Kun Ouyang, Liang Chen, Lin Sui, Longhui Yu, Mengfan Dong, Mengnan Dong, Nuo Xu, Pengyu Cheng, Qizheng Gu, Runjie Zhou, Shaowei Liu, Sihan Cao, Tao Yu, Tianhui Song, Tongtong Bai, Wei Song, Weiran He, Weixiao Huang, Weixin Xu, Xiaokun Yuan, Xingcheng Yao, Xingzhe Wu, Xinxing Zu, Xinyu Zhou, Xinyuan Wang, Y. Charles, Yan Zhong, Yang Li, Yangyang Hu, Yanru Chen, Yejie Wang, Yibo Liu, Yibo Miao, Yidao Qin, Yimin Chen, Yiping Bao, Yiqin Wang, Yongsheng Kang, Yuanxin Liu, Yulun Du, Yuxin Wu, Yuzhi Wang, Yuzi Yan, Zaida Zhou, Zhaowei Li, Zhejun
  Jiang, Zheng Zhang, Zhilin Yang, Zhiqi Huang, Zihao Huang, Zijia Zhao, and Ziwei Chen.
\newblock {Kimi-VL} technical report, 2025.

\bibitem[Wang et~al.(2024{\natexlab{a}})Wang, Bai, Tan, Wang, Fan, Bai, Chen, Liu, Wang, Ge, Fan, Dang, Du, Ren, Men, Liu, Zhou, Zhou, and Lin]{Qwen2VL}
Peng Wang, Shuai Bai, Sinan Tan, Shijie Wang, Zhihao Fan, Jinze Bai, Keqin Chen, Xuejing Liu, Jialin Wang, Wenbin Ge, Yang Fan, Kai Dang, Mengfei Du, Xuancheng Ren, Rui Men, Dayiheng Liu, Chang Zhou, Jingren Zhou, and Junyang Lin.
\newblock Qwen2-vl: Enhancing vision-language model's perception of the world at any resolution.
\newblock \emph{arXiv preprint arXiv:2409.12191}, 2024{\natexlab{a}}.

\bibitem[Wang et~al.(2025)Wang, Wang, Suzuki, and Kobayashi]{wang2025seeing}
Wei-Yao Wang, Zhao Wang, Helen Suzuki, and Yoshiyuki Kobayashi.
\newblock Seeing is understanding: Unlocking causal attention into modality-mutual attention for multimodal llms.
\newblock \emph{arXiv preprint arXiv:2503.02597}, 2025.

\bibitem[Wang et~al.(2021)Wang, Qin, and Cui]{wang2021occlusion}
Yiding Wang, Yuxin Qin, and Jiali Cui.
\newblock Occlusion robust wheat ear counting algorithm based on deep learning.
\newblock \emph{Frontiers in Plant Science}, 12:\penalty0 645899, 2021.

\bibitem[Wang et~al.(2024{\natexlab{b}})Wang, Wang, and Huang]{wang2024dual}
Yongjie Wang, Feng Wang, and Dongyang Huang.
\newblock Dual-branch counting method for dense crowd based on self-attention mechanism.
\newblock \emph{Expert Systems with Applications}, 236:\penalty0 121272, 2024{\natexlab{b}}.

\bibitem[Wynn(1990)]{wynn1990children}
Karen Wynn.
\newblock Children's understanding of counting.
\newblock \emph{Cognition}, 36\penalty0 (2):\penalty0 155--193, 1990.

\bibitem[Xu et~al.(2024)Xu, Zhang, and Shi]{xu2024amodal}
Katherine Xu, Lingzhi Zhang, and Jianbo Shi.
\newblock Amodal completion via progressive mixed context diffusion.
\newblock In \emph{Proceedings of the IEEE/CVF Conference on Computer Vision and Pattern Recognition}, pages 9099--9109, 2024.

\bibitem[Yao et~al.(2024)Yao, Yu, Zhang, Wang, Cui, Zhu, Cai, Li, Zhao, He, et~al.]{yao2024minicpm}
Yuan Yao, Tianyu Yu, Ao Zhang, Chongyi Wang, Junbo Cui, Hongji Zhu, Tianchi Cai, Haoyu Li, Weilin Zhao, Zhihui He, et~al.
\newblock Minicpm-v: A gpt-4v level mllm on your phone.
\newblock \emph{arXiv preprint arXiv:2408.01800}, 2024.

\bibitem[Zhou et~al.(2024)Zhou, Rao, Li, Hu, and Sun]{zhou2024multi}
Lifang Zhou, Songlin Rao, Weisheng Li, Bo Hu, and Bo Sun.
\newblock Multi-branch progressive embedding network for crowd counting.
\newblock \emph{Image and Vision Computing}, page 105140, 2024.

\end{thebibliography}
}
\newpage
\newpage
\appendix

\section*{Appendix}

\section{Implementation Details}

\subsection{Metric Details}
\label{appendix:metric}
We use symmetric mean percent error (sMAPE) as the primary metric for our benchmarks due to its resistance to bias for under/over predictions and small/large ground truths \citep{maiseli2019optimum}. 
The standard metric for a counting benchmark is mean average error (MAE).
MAE is popular, but heavily penalizes predictions that deviate by a small margin from big ground truths, highlighting the necessity for a metric that gives equal weighting to all questions. 
Mean average percent error (MAPE) initially seems appealing
but is disproportionally inflated for small ground truths and is biased towards overpredictions. Mean square error (MSE) and root mean square error (RMSE) are also commonly used but are very sensitive to outliers because they square the error. Intuitively, performing well on almost all questions and poorly on a small subset should score better than consistently being wrong. Among commonly-used metrics, sMAPE is the only metric that evaluates performance in relation to the distribution of ground truth elements \citep{chicco2021coefficient}. There are two common definitions \citep{flores1986pragmatic} for sMAPE, but we use the one that scales to 100\%.
sMAPE is given by:

\begin{equation}
\text{sMAPE} = 100 \cdot \frac{1}{n} \sum_{i=1}^{n} \frac{|y_i - \hat{y}_i|}{|y_i| + |\hat{y}_i|}
\end{equation}

\noindent where \( y_i \) represents the actual values, \( \hat{y}_i \) represents the predicted values, and \( n \) is the number of observations. sMAPE is capped at 100\%, providing a finite scoring range. This feature is ideal for challenging tasks like ours, as it penalizes model responses that fail to produce an answer.

\subsection{Output Tokens}
\label{appendix:output_token}
To maximize the VLM's chance at success, we allocate a high number of output tokens to generate a rationale and output. 
This varies per model.
We give 4000 tokens to InternVL2, 2000 tokens to Molmo, and 8192 tokens to Qwen2VL, following their max output lengths.
For GPT-4o, we use the default of 4096 tokens. 
\section{\method{} Dataset Creation Details}
\label{appendix:dataset}

The following expands upon \cref{sec:dataset}.
While FSC-147, a diverse counting dataset with manual annotations, is a strong starting point, it cannot immediately be adapted to our task. 
To make the task of amodal counting solvable, our dataset requires images with patterns in them. A person (or model) can infer how the pattern would continue and thus accurately predict the total number. 
For questions to be answerable, the dataset's images must be filtered down to represent patterns a model or person could recognize.

Our filtering process follows two stages. 
\textbf{First}, we prompt GPT-4o to determine whether the objects were arranged in a pattern. 
\textbf{Second}, if the model responded with ``no'', the images were immediately discarded.
If the model output was ``yes'', the log probability of the token is stored. 
Empirically, we found that higher log probability values (i.e. higher confidence scores) corresponded to more well-defined patterns in the image. 
Thus, we use the log probabilities for filtering.

Specifically, let \( P_{\text{yes}} \) be the log probability of the ``yes'' token and \( T \) denote the threshold for determining how well-defined a pattern is.\footnote{We set \( T = 0.9999 \) based on manual evaluation, finding it resulted in fewer false positives.}
To filter the images based on pattern rigidity, we apply the following condition: $e^{P_{\text{yes}}} \geq T$.
This inequality yields 991 images from the original dataset (16.12\%). 
Next, we manually filter each of the selected images to ensure that they indeed contain patterns and feature a countable number of objects, excluding 34 images.  
Afterward, we manually place a ``fair'' occluding box in each image, i.e. a box that leaves sufficient portions of the pattern visible, such that the pattern can still be inferred from the unoccluded portions of the image. 
Occluding boxes were also chosen with varying positions and sizes in the image. 

\section{Additional Analysis}
Here we provide additional experiments that attempt to either increase model performance on \method{} or dissect the reasons behind poor model performance. Chain-of-Thought inhibits model performance, while temperature backoff slightly improves performance. Additionally, we find that models struggle at counting just occluded objects, are overconfident in occluded settings, and are biased to predict specific numbers. 

\subsection{Chain-of-Thought reduces model performance}

\begin{table}[h]
\centering
\resizebox{.8\columnwidth}{!}{
\begin{tabular}{lcc}
\toprule
Method & \real{} & \synthetic{} \\
\midrule
GPT-4o & 14.75 & 9.71\\
GPT-4o w/ CoT & 14.94 & 7.73\\
Qwen2 & 29.33 & 11.74\\
Qwen2 w/ CoT & 31.57 & 37.81\\
\bottomrule
\end{tabular}
}
\vspace{-0.75em}
\caption{CoT experiments (metric: sMAPE).}
\label{tab:cot}
\end{table}
\noindent
During development, we experimented with several common strategies including CoT.
In \cref{tab:cot}, we find that CoT reduces model performance except in the occluded synthetic scenario, most likely because  the included examples are very similar to the test prompt.

\subsection{Temperature backoff slightly improves model performance} 

\begin{table*}[h]
\centering
\resizebox{\textwidth}{!}{%
\begin{tabular}{lcccccccc}
\toprule
\multirow{3}{*}{Model} 
& \multicolumn{8}{c}{Error (\%) ($\downarrow$)} \\
\cmidrule(lr){2-9}
& \multicolumn{4}{c}{Real} 
& \multicolumn{4}{c}{Synthetic} \\
\cmidrule(lr){2-5} \cmidrule(lr){6-9}
& \multicolumn{2}{c}{Unoccluded} & \multicolumn{2}{c}{Occluded} 
& \multicolumn{2}{c}{Unoccluded} & \multicolumn{2}{c}{Occluded} \\
\cmidrule(lr){2-3} \cmidrule(lr){4-5} \cmidrule(lr){6-7} \cmidrule(lr){8-9}
& Original & w/ backoff (\(\Delta\)) 
& Original & w/ backoff (\(\Delta\)) 
& Original & w/ backoff (\(\Delta\)) 
& Original & w/ backoff (\(\Delta\)) \\\midrule
GPT-4o
& 13.34 & 12.57 (\(-0.77\)) 
& 14.75 & 14.39 (\(-0.36\)) 
&  5.90 &  5.93 (\(+0.03\)) 
&  9.71 &  9.23 (\(-0.48\)) \\
InternVL2
& 26.17 & 27.09 (\(+0.92\)) 
& 32.90 & 32.37 (\(-0.53\)) 
& 16.44 & 15.59 (\(-0.85\)) 
& 17.57 & 16.24 (\(-1.33\)) \\
Molmo
& 25.90 & 21.23 (\(-4.67\)) 
& 32.49 & 28.17 (\(-4.32\)) 
&  8.40 &  2.88 (\(-5.52\)) 
& 17.73 & 15.85 (\(-1.88\)) \\
Qwen2VL
& 18.96 & 19.40 (\(+0.44\)) 
& 29.33 & 28.47 (\(-0.86\)) 
&  6.63 &  6.66 (\(+0.03\)) 
& 11.74 & 11.51 (\(-0.23\)) \\
\midrule
Avg.\ of 4 VLMs
& 21.09 & 20.07 (\(-1.02\)) 
& 27.37 & 25.85 (\(-1.52\)) 
&  9.34 &  7.76 (\(-1.58\)) 
& 14.19 & 13.21 (\(-0.98\)) \\
\bottomrule
\end{tabular}
}
\caption{
Comparison of models on \method{} across four scenarios (\real{} vs.\ \synthetic{}, Unoccluded vs.\ Occluded). 
“Original” indicates no backoff; “w/ backoff” indicates applying backoff, with \(\Delta\) = (\textit{w/ backoff}) \(-\) (\textit{Original}). 
Negative \(\Delta\) values indicate an improvement.
}
\label{tab:backoff}
\end{table*}

To improve VLM performance on \method{}, we address a trend we established during early testing.
Most of the time, the VLM fails by reaching an incorrect answer.
Sometimes, however, our benchmark can cause VLMs to produce a long and irrelevant response that strays from the original prompt, leading to the worst possible sMAPE score (100\%).

To reduce the number of skipped questions, we experiment with \emph{temperature backoff}, which iteratively decreases the sampling temperature. 
Because the answer extractor can immediately identify an incoherent output, we can regenerate the response with a lower temperature to get the model to answer the task properly.
Consistent with our findings, \citet{peeperkorn2024temperature} also finds that lower temperatures increase coherence in VLMs, thereby enhancing their chances of maintaining relevance to the prompt. 
Therefore, temperature backoff gives VLMs a better chance of achieving higher scores. 
Each time the answer extractor returns an empty answer because the VLMs produced an incoherent answer, we reduce the temperature by $0.1$ (starting from $1.0$) until it reaches $0.0$, at which point the example is skipped.

\myparagraph{Models perform slightly better with temperature backoff.}
We introduced temperature backoff to reduce model incoherence, and it performed fairly well.
As shown in \cref{tab:backoff} (bottom), this method slightly improves performance across each model, resulting in an average error reduction of 5.78\% in \real{} and 5.45\% in \synthetic{}.
Temperature backoff essentially allows the model to reattempt the question if it fails to respond to the prompt.
Similar to previous results, positive results from reattempts highlight VLMs' weak reasoning abilities.

\subsection{Models struggle at counting just occluded objects}
\label{appendix:only_occ}

We separately test whether models can count only the occluded objects (not including the visible objects) in an image. 
Here, as \cref{tab:only_occluded} demonstrates, the models perform especially poorly in this task, with high error rates across all models. 
Therefore, we can conclude that occlusion and counting are uniquely difficult for the VLMs, and that the drop in performance between unoccluded and occluded settings in \cref{tab:real_unocc_occ} is likely due to a poor ability to count occluded objects. 

\begin{table}[t]
\centering
\begin{tabular}{lcc}
\toprule
\multirow{2}{*}{Model} & \multicolumn{2}{c}{Error (\%) [$\downarrow$]}\\
\cmidrule(lr){2-3}
 & All Objects & Only Occluded\\
\midrule
GPT-4o & 14.75 & 26.13 (+11.38)\\
InternVL2 & 32.90 & 75.82 (+42.92)\\
Molmo & 32.49 & 96.79 (+64.30)\\
Qwen2VL & 29.33 & 32.89 (+3.56)\\
\midrule
Avg. of 4 VLMs & 27.37 & 57.91 (+30.54)\\
\bottomrule
\end{tabular}
\caption{
VLM sMAPE for counting all objects and counting only the occluded objects in \real{}. Metric: sMAPE (lower is better).}
\label{tab:only_occluded}
\end{table}
\subsection{Models are overconfident in occluded settings}
\begin{figure}[h]
    \centering
    \includegraphics[width=\linewidth]{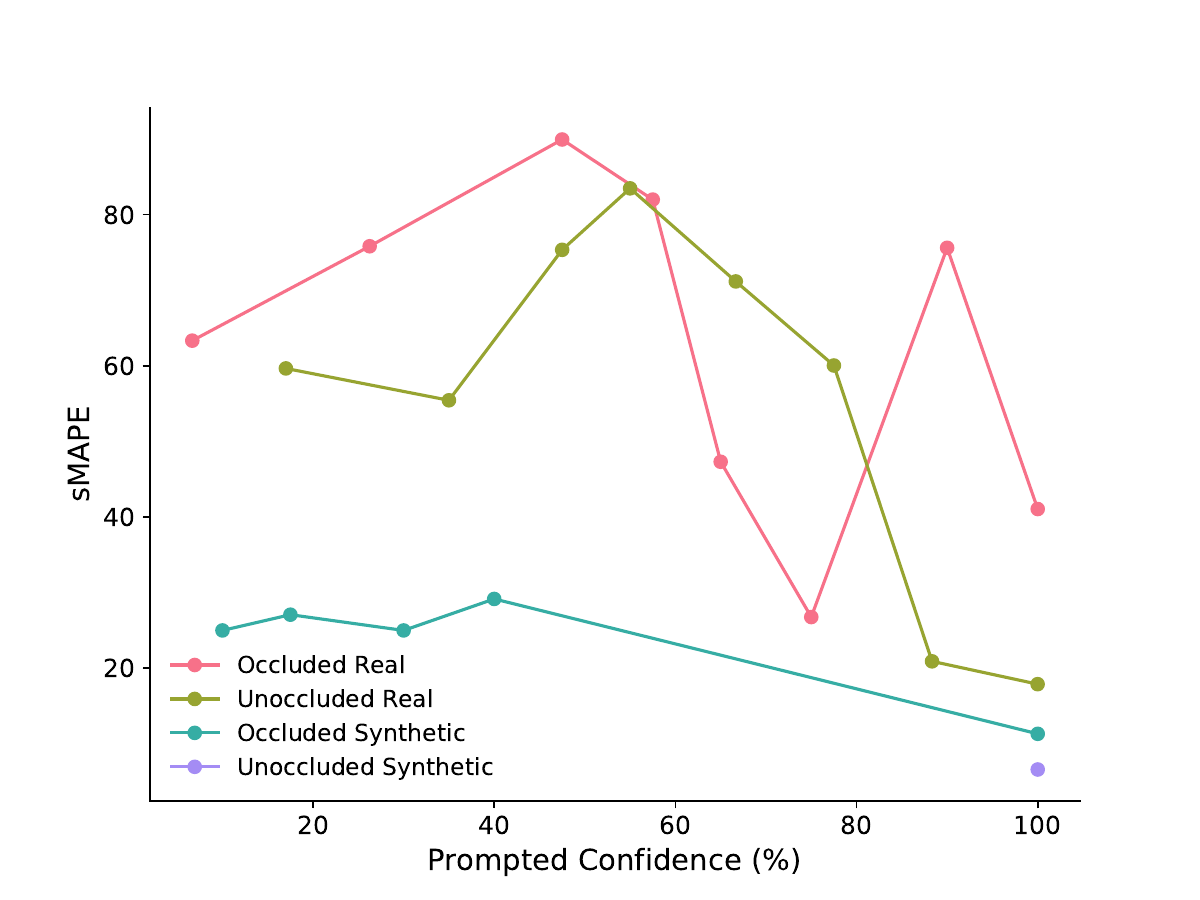}
    \caption{Reliability curve of prompting model for confidence vs. sMAPE.}
    \label{fig:prompted_uncertainty}
\end{figure}

\begin{figure}[h]
    \centering
    \includegraphics[width=\linewidth]{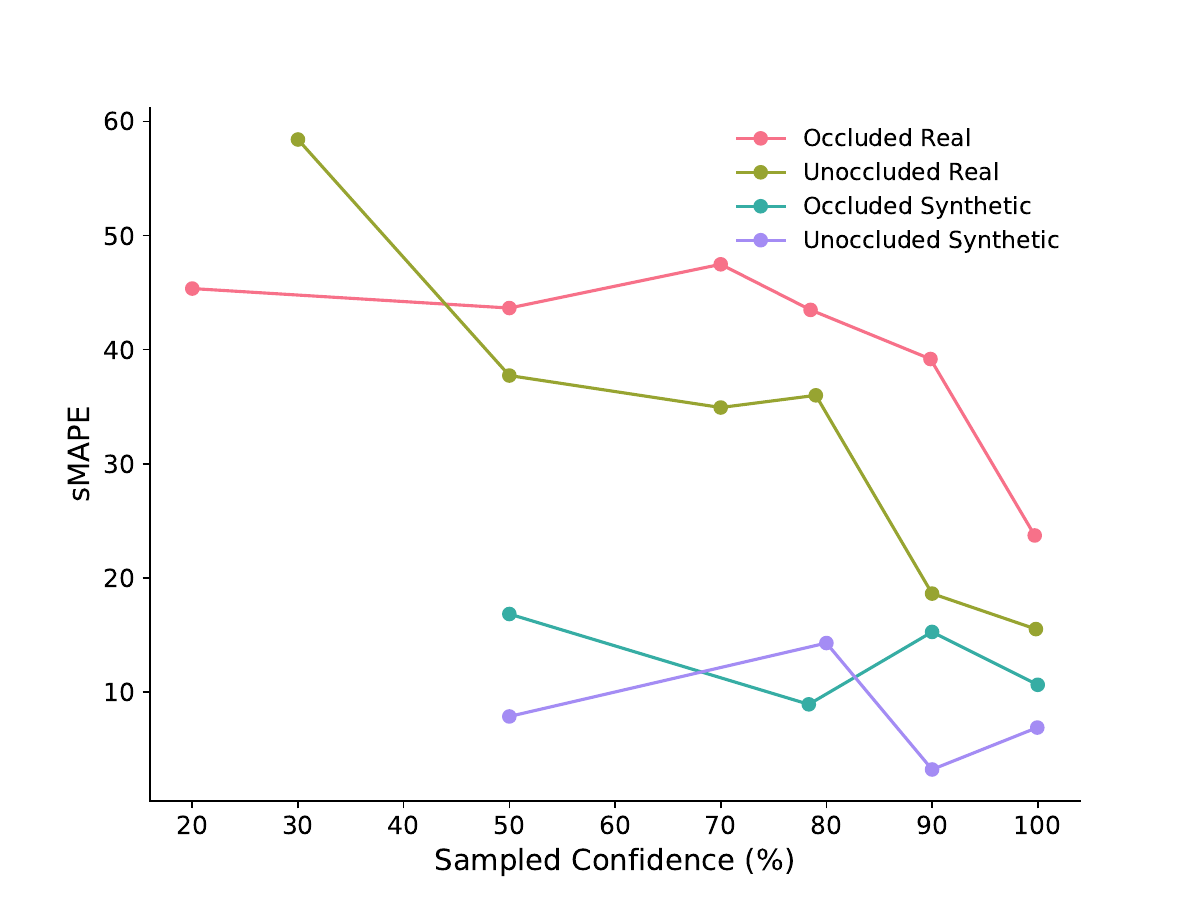}
    \caption{Reliability curve of sampling model for confidence vs. sMAPE.}
    \label{fig:sampled_uncertainty}
\end{figure}

We test the uncertainty with two different methods of obtaining confidence on Qwen2VL. In the first method, we prompt Qwen2VL for its confidence in the answer. For the second method, we generate 20 responses for every question in our VQA and calculate the confidence as the percentage of times the most common answer was generated. These results can be seen in \cref{fig:prompted_uncertainty} and \cref{fig:sampled_uncertainty} 
respectively. In both reliability curves, there is a slight trend that the model's confidence is negatively correlated with the error, which is the desired outcome. 
In \real{}, however, the correlation is much stronger. 
While the models are somewhat calibrated (with generally lower confidence on higher-error examples, there are still outliers in prompted confidence for \real{} occluded and sampled confidence for \synthetic{} occluded. 
This indicates that not only do the models perform worse under occlusion, but they can also be overconfident.

\subsection{Models are biased to predict specific numbers.} \label{appendix:heatmap}

To examine where models frequently err, we generated a confusion matrix for every model based on \synthetic{} results (shown in \cref{appendix:heatmap}). 
The y-axis represents the ground truth values and the x-axis represents the model's answers.
We find that models often over-predict numbers associated with common counts in real life: 
GPT-4o tends to predict numbers like 8, 9, 10, and 12, which are all non-prime numbers (i.e. can be arranged into a grid) and common  groupings of objects. For example, 12 is a common grouping (dozens) and allows arrangements into 3x4 or 2x6 grids.
InternVL and Qwen2VL over-predict 5 and 10, aligning with how humans conceptualize numbers. Indeed, \citet{coupland2011frequent} found that numbers 5, 10, 20, and other round numbers appear disproportionally more in online texts.
Molmo has no correlation with these factors, possibly due to its unique ``point and count'' ability.

\begin{figure*}[h]
    \centering
    \includegraphics[width=\linewidth]{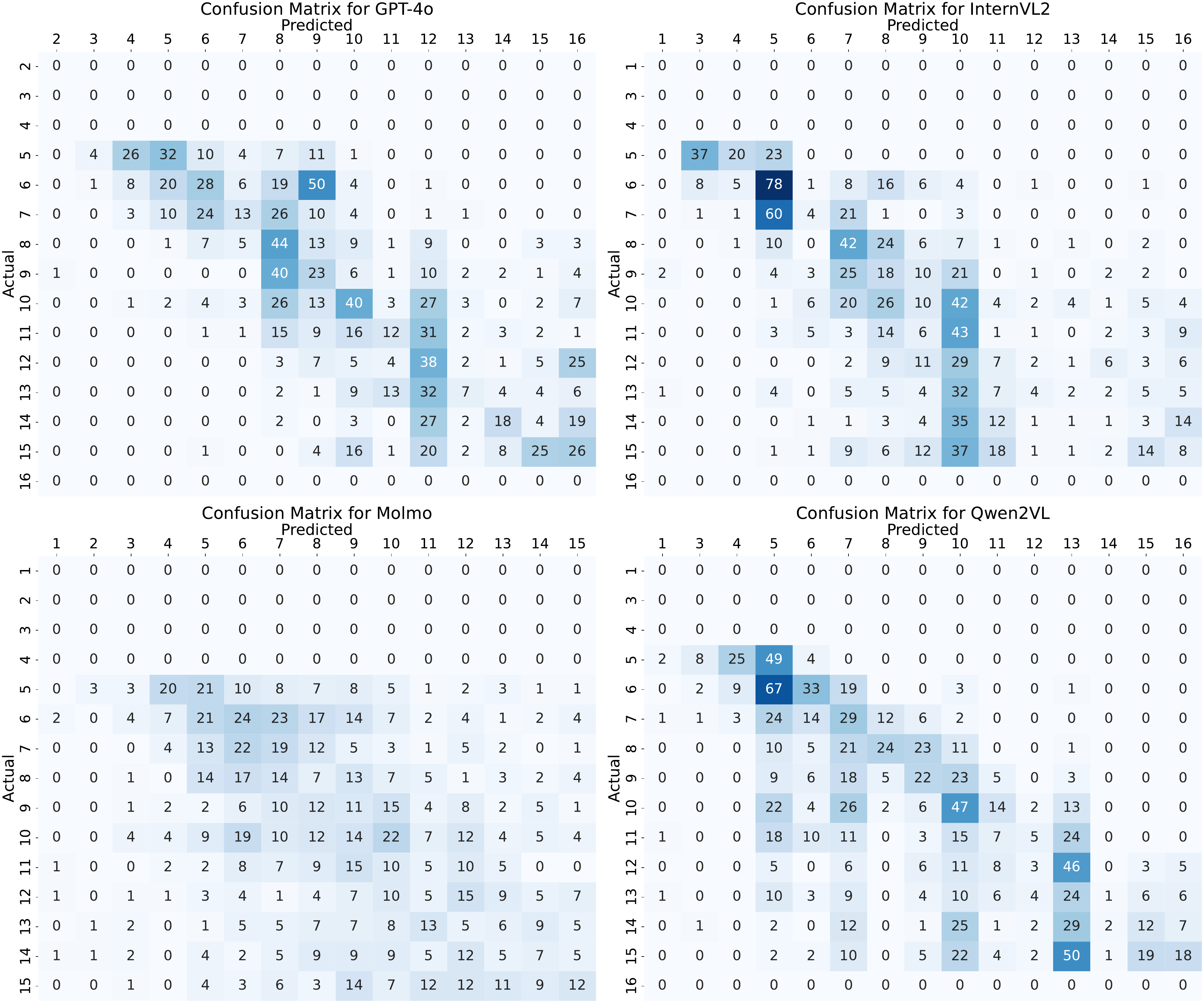}
    \caption{Confusion matrix: predicted vs. ground truth counts for \real{}'s occluded split.}
    \label{fig:heatmap}
\end{figure*}

\section{VLM Prompts}
\label{appendix:prompts}
We use a 100-example validation set for each setting to select the best prompt, which we report below.

\begin{user_example}[frametitle={Prompt for GPT-4o on \real{} unoccluded split.}]
Count the exact number of [object] in the image. Assume the pattern of [object] continues behind any black box. Provide the total number of [object] as if the black box were not there.
\end{user_example}

\begin{user_example}[frametitle={Prompt for InternVL2 on \real{} unoccluded split.}]
Your task is to count objects in the image. First, state what the pattern is, then give your final count.
\end{user_example}

\begin{user_example}[frametitle={Prompt for Molmo on \real{} unoccluded split.}]
Count the exact number of [object] in the image. Only count [object] that are visible within the frame. If [object] are partially in the frame (i.e. if any part of [object] are visible), count it.
\end{user_example}

\begin{user_example}[frametitle={Prompt for Qwen2VL on \real{} unoccluded split.}]
Count the exact number of [object] in the image. Assume the pattern of [object] continues behind any black box. Provide the total number of [object] as if the black box were not there. Only count [object] that are visible within the frame (or would be visible without the occluding box). If [object] are partially in the frame (i.e. if any part of [object] are visible), count it. If the [object] would be partially in the frame without the occluding box, count it.
\end{user_example}

\begin{user_example}[frametitle={Prompt for GPT-4o, InternVL2, and Qwen2VL on \real{} occluded split.}]
Count the exact number of [object] in the image. Assume the pattern of [object] continues behind any black box. Provide the total number of [object] as if the black box were not there. Only count [object] that are visible within the frame (or would be visible without the occluding box). If [object] are partially in the frame (i.e. if any part of [object] are visible), count it. If the [object] would be partially in the frame without the occluding box, count it.
Molmo: Your task is to count objects in the image. Assume the pattern of [object] continues behind the black box. First, state what the pattern is, then give your final count.
\end{user_example}

\begin{user_example}[frametitle={Prompt for Molmo on \real{} occluded split.}]
Your task is to count objects in the image. Assume the pattern of [object] continues behind the black box. First, state what the pattern is, then give your final count.
\end{user_example}

\begin{user_example}[frametitle={Prompt for GPT-4o on \synthetic{} unoccluded split.}]
Your task is to count objects in the image. First, state what the pattern is, then give your final count.
\end{user_example}

\begin{user_example}[frametitle={Prompt for InternVL2 on \synthetic{} unoccluded split.}]
Count the exact number of [dot shape]s in the image. Only count [dot shape]s that are visible within the frame. If [dot shape]s are partially in the frame (i.e. if any part of [dot shape]s are visible), count it.
\end{user_example}

\begin{user_example}[frametitle={Prompt for Molmo on \synthetic{} unoccluded split.}]
Count the exact number of [dot shape]s in the image. Only count [dot shape]s that are visible within the frame.
\end{user_example}

\begin{user_example}[frametitle={Prompt for Qwen2VL on \synthetic{} unoccluded split.}]
Count the exact number of [dot shape]s in the image. Assume the pattern of [dot shape]s continues behind any black box. Provide the total number of [dot shape]s as if the black box were not there. Only count [dot shape]s that are visible within the frame (or would be visible without the occluding box). If [dot shape]s are partially in the frame (i.e. if any part of [dot shape]s are visible), count it. If the [dot shape]s would be partially in the frame without the occluding box, count it.
\end{user_example}

\begin{user_example}[frametitle={Prompt for GPT-4o and Molmo on \synthetic{} occluded split.}]
Your task is to count objects in the image. Assume the pattern of [dot shape]s continues behind the black box. First, state what the pattern is, then give your final count.
\end{user_example}

\begin{user_example}[frametitle={Prompt for InternVL2 and Qwen2VL on \synthetic{} occluded split.}]
Count the exact number of [dot shape]s in the image. Assume the pattern of [dot shape]s continues behind any black box. Provide the total number of [dot shape]s as if the black box were not there. Only count [dot shape]s that are visible within the frame (or would be visible without the occluding box). If [dot shape]s are partially in the frame (i.e. if any part of [dot shape]s are visible), count it. If the [dot shape]s would be partially in the frame without the occluding box, count it.
\end{user_example}

\end{document}